\documentclass{article}

\usepackage[final]{neurips_2020}
\usepackage[utf8]{inputenc} 
\usepackage[T1]{fontenc}    
\usepackage{hyperref}       
\usepackage{url}            
\usepackage{booktabs}       
\usepackage{amsfonts}       
\usepackage{nicefrac}       
\usepackage{microtype}      
\usepackage[toc,page]{appendix}

\usepackage{amsmath}
\usepackage{graphicx}
\usepackage{natbib}
\usepackage{wrapfig}
\usepackage{subfig}
\usepackage{amsthm}
\newtheorem{theorem}{Theorem}[section]

\newtheorem{lemma}[theorem]{Lemma}

\usepackage{mathtools}
\DeclarePairedDelimiter\abs{\lvert}{\rvert}

\newtheorem{innercustomgeneric}{\customgenericname}
\providecommand{\customgenericname}{}
\newcommand{\newcustomtheorem}[2]{%
  \newenvironment{#1}[1]
  {%
   \renewcommand\customgenericname{#2}%
   \renewcommand\theinnercustomgeneric{##1}%
   \innercustomgeneric
  }
  {\endinnercustomgeneric}
}

\newcustomtheorem{customthm}{Theorem}
\newcustomtheorem{customlemma}{Lemma}
\usepackage[usenames,dvipsnames]{xcolor}
\newcommand{\beginsupplement}{%
        \setcounter{table}{0}
        \renewcommand{\thetable}{S\arabic{table}}%
        \setcounter{figure}{0}
        \renewcommand{\thefigure}{S\arabic{figure}}%
     }

\title{Can I Trust My Fairness Metric? Assessing Fairness with Unlabeled Data and Bayesian Inference
} 

\author{%
  Disi Ji$^1$ \hspace{1em}
  \textbf{Padhraic Smyth}$^1$ \hspace{1em}
  \textbf{Mark Steyvers}$^2$\\
  {}$^1$\text{Department of Computer Science}\hspace{1em}
  {}$^2$\text{Department of Cognitive Sciences}\\
  University of California, Irvine\\
  \texttt{disij@uci.edu} \hspace{1em} 
  \texttt{smyth@ics.uci.edu} \hspace{1em} 
  \texttt{mark.steyvers@uci.edu} \\
}

\begin{document}
\maketitle

\begin{abstract}
We investigate the problem of reliably assessing group fairness when labeled examples are few  but unlabeled examples are plentiful.  
We propose a general Bayesian framework  that can augment labeled data with unlabeled data to produce more accurate and lower-variance estimates compared to methods based on labeled data alone. Our approach estimates calibrated scores for unlabeled examples in each group using a hierarchical latent variable model conditioned on labeled examples. 
This in turn allows for inference of posterior distributions with associated notions of uncertainty for a variety of group fairness metrics.
We demonstrate that our approach leads to significant and consistent reductions in estimation error across multiple well-known fairness datasets, sensitive attributes, and predictive models. 
The results show the benefits of using both unlabeled data and Bayesian inference in terms of assessing whether a prediction model is fair or not.
\end{abstract}

\section{Introduction}
 
Machine learning models are increasingly used to make important decisions about individuals. At the same time it has become apparent that these models are susceptible to producing systematically biased decisions with respect to sensitive attributes such as gender, ethnicity, and age~\citep{angwin2017we, berk2018fairness, corbett2018measure, chen2019can, beutel2019putting}.
This has led to a significant amount of recent work in machine learning addressing these issues,  including research on both (i) definitions of fairness in a machine learning context (e.g.,~\cite{dwork2012fairness, chouldechova2017fair}), and (ii) design of fairness-aware learning algorithms that can mitigate issues such as algorithmic bias (e.g.,~\cite{calders2010three, kamishima2012fairness, feldman2015certifying, zafar2017fairness, chzhen2019leveraging}). 

In this paper we focus on an important yet under-studied aspect of the fairness problem:   reliably assessing how fair a blackbox model is, given limited labeled data. In particular, we focus on assessment of group fairness of binary classifiers.
Group fairness is measured with respect to parity in prediction performance between different demographic groups. Examples include differences in performance for metrics such as true positive rates and false positive rates (also known as equalized odds~\citep{hardt2016equality}), accuracy~\citep{chouldechova2017fair}, false discovery/omission rates~\citep{zafar2017fairness},   and calibration and balance~\citep{kleinberg2016inherent}.

\begin{figure}[t]
\centering 
   \subfloat{\includegraphics[width=\linewidth]{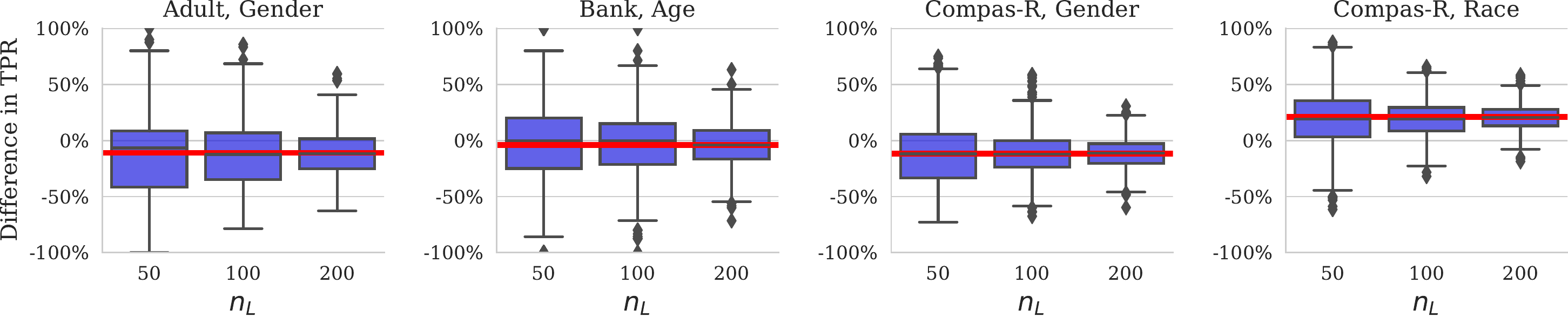}}
  \caption{Boxplots of frequency-based estimates of the difference in true positive rate (TPR) for four fairness datasets and binary sensitive attributes, across 1000 randomly sampled sets of labeled test examples of size $n_L=50,100,200$. The horizontal red line is the TPR difference computed on the full test dataset.} 
  \label{fig:freqerror}
\end{figure}

Despite the simplicity of these definitions, a significant challenge in assessment of group fairness is high variance in estimates of these metrics based on small amounts of labeled data. 
To illustrate this point,
Figure \ref{fig:freqerror} shows frequency-based estimates of group differences in true positive rates (TPRs) for four real-world datasets. The boxplots clearly show the high variability for the estimated TPRs relative to the true TPRs (shown in red) as a function of the number of labeled examples $n_L$. In many cases the estimates are two or three or more times larger than the true difference. In addition, a relatively large percentage of the estimates have the opposite sign of the true difference, potentially leading to mistaken conclusions. 
The variance of these estimates decreases relatively slowly, e.g., at a rate of approximately $\frac{1}{n}$ for group differences in accuracy where $n$ is the number of labels in the smaller of the two groups\footnote{Stratified sampling by group could help with this issue (e.g., see \cite{sawade2010}), but stratification might not always be possible in practice, and the total variance will still converge slowly overall.}. 
Imbalances in label distributions can further compound the problem, for example for estimation of group differences in TPR or FPR.
For example, consider a simple simulation with two groups, where the underrepresented group makes up 20\% of the whole dataset, groupwise positive rates $P(y=1)$ are both 20\%, and the true groupwise TPRs are 95\% and  90\% (additional details in the Supplement). In order to ensure that there is a 95\% chance that our estimate of the true TPR difference (which is 0.05) lies in the range [0.04, 0.06] we need at least 96k labeled instances.
Yet for real-world datasets used in the fairness literature (e.g., \cite{friedler2019comparative}; see also Table~\ref{tab:data} later in the paper),  test set sizes  are often much smaller than this, and it is not uncommon for the group and label distributions to be even more imbalanced. 

The real-world and synthetic examples above show that frequentist assessment of group fairness is unreliable  unless the labeled dataset is unrealistically large. 
Acquiring large amounts of labeled data can be difficult and time-consuming, particularly for the types of applications where fairness is important,  such as decision-making in medical or criminal justice contexts~\citep{angwin2017we,berk2018fairness,rajkomar2018ensuring}. This is in contrast to applications such as image classification where approaches like Mechanical Turk can be used to readily generate large amounts of labeled data.

To address this problem, we propose to augment labeled data with unlabeled data to generate more accurate and lower-variance estimates compared to methods based on labeled data alone. 
In particular, the three primary contributions of this paper are (1) a comprehensive Bayesian treatment of fairness assessment that provides uncertainty about estimates of group fairness metrics; (2) a new Bayesian methodology that uses calibration to leverage information from both unlabeled and labeled examples; and, (3) systematic large-scale experiments across multiple datasets, models, and attributes that show that using unlabeled data can reduce  estimation error significantly.

\section{Fairness Assessment with Bayesian Inference and Unlabeled Data} 

\subsection{Notation and Problem Statement}
Consider a trained binary classification model $M$, with inputs $x$ and class labels $y \in \{0,1\}$.
The model produces scores\footnote{Note that the term ``score" is sometimes defined differently in the calibration literature as the maximum class probability for the model. Both definitions are equivalent mathematically for binary classification.}
$s  = P_M( y  = 1 | x ) \in [0, 1]$, where  $P_M$ denotes the fact that this is the model's estimate of the probability that $y = 1$ conditioned on $x$. Under 0-1 loss the model predicts $\hat{y}=1$ if $s \ge 0.5$ and $\hat{y}=0$ otherwise. The marginal accuracy of the classifier is $P(\hat{y} = y)$ and the accuracy conditioned on a particular value of the score $s$ is  $P(\hat{y} = y | s)$. 
A classifier is calibrated if $P(\hat{y} = y) | s) = s$, e.g., if  whenever the model produces a score of $s=0.9$  then its prediction is correct 90\% of the time.
For group fairness we are interested in potential  differences in performance metrics with respect to a sensitive attribute (such as gender or race) whose values $g$ correspond to different groups, $g \in \{0, 1, \ldots, G-1\}$. We will use $\theta_g$ to denote a particular metric of interest, such as accuracy, TPR, FPR, etc. for group $g$.  We focus on group differences for two groups, defined as $\Delta  = \theta_1 - \theta_0$, e.g., the difference in a model's predictive accuracy between females and males, $\Delta =P(\hat{y} = y | g=1) - P(\hat{y} = y| g=0)$.

We assume in general that the available data consists of both $n_L$ labeled examples and $n_U$ unlabeled examples, where $n_L \ll n_U$, which is a common situation in practice where far more unlabeled data is available than labeled data.  For the unlabeled examples, we do not have access to the true labels $y_j$ but we do have the scores $s_j = P_M( y_j = 1 | x_j)$, $j=1,\ldots,n_U$. For the labeled examples, we have the true labels $y_i$ as well as the scores $s_i$, $i=1,\ldots,n_L$.  The examples (inputs $x$, scores $s$, and labels $y$ if available) are sampled IID from an underlying joint distribution $P(x,y)$ (or equivalently $P(s,y)$ given that $s$ is a deterministic function via $M$ of $x$), where this underlying distribution represents the population we wish to evaluate fairness with respect to. Note that in practice $P(x,y)$  might very well not be the same distribution the model was trained on. For unlabeled data $D_u$ the corresponding distributions are $P(x)$ or $P(s)$. 

\subsection{Beta-Binomial Estimation with Labeled Data}
Consider initially the case with only labeled data $D_L$ (i.e., $n_U = 0$) and for simplicity let the metric of interest $\Delta$ be group difference in classification accuracy. Let $I_i=I_{\hat{y}_i=y_i}, 1 \le i \le n_L$, be a binary indicator of whether each labeled example $i$ was classified correctly or not by the model. 
The binomial likelihood for  group accuracy $\theta_g, g=0,1$, treats the $I_i$'s as conditionally independent draws from a true unknown accuracy $\theta_g$, $I_i \sim \mathrm{Bernoulli}( \theta_g)$.
We can perform  Bayesian inference on the $\theta_g$'s   by specifying conjugate $\mathrm{Beta}(\alpha_g, \beta_g)$ priors for each $\theta_g$, combining these priors with the binomial likelihoods, and obtaining posterior densities in the form of the beta densities on each $\theta_g$.  
From here we can get a posterior density on the group difference in accuracy, $P(\Delta | D_L)$ where $\Delta = \theta_1 - \theta_0$. Since the density for the difference of two beta-distributed quantities (the $\theta$'s)  is not in general in closed form, we  use posterior simulation (e.g.,~\cite{gelman2013bayesian}) to obtain posterior samples of $\Delta$ by sampling $\theta$'s from their posterior densities and taking the difference.   For metrics such as TPR we place beta priors on  conditional quantities such as $\theta_g=P(\hat{y}=1 | y=1, g)$. In all of the results in the paper we use weak uninformative priors for $\theta_g$ with $\alpha_g = \beta_g = 1$. This general idea of using Bayesian inference on classifier-related metrics  has been noted before for metrics such  marginal accuracy~\citep{benavoli2017time}, TPR~\citep{johnson2019gold}, and precision-recall~\citep{goutte2005probabilistic}, but has not been developed or evaluated in the context of fairness assessment.
  
This beta-binomial approach above provides a useful, simple, and practical tool for understanding and visualizing   uncertainty about fairness-related metrics, conditioned on a set of $n_L$ labeled examples. 
However, with weak uninformative priors, the posterior density for $\Delta$ will  be relatively wide unless $n_L$ is very large,  analogous to the high empirical variance  for frequentist point estimates in Figure 1. 
As with the frequentist variance, the width of the posterior density on $\Delta$  will decrease relatively slowly at a rate of approximately $\frac{1}{n_L}$. 
This motivates the main goal of the paper: 
can we combine unlabeled examples with labeled examples to make more accurate inferences about fairness metrics?

\subsection{Leveraging Unlabeled Data with a Bayesian Calibration Model}
Consider the situation where we have $n_U$ unlabeled examples, in addition to the $n_L$ labeled ones.
For each unlabeled example $j = 1,\ldots,n_U$ we can use the model $M$ to generate a score, $s_j = P_M( y_j = 1 | x_j)$. If the model $M$ is perfectly calibrated then the model's score is the true probability that $y=1$, i.e., we have $s_j = P_M( y_j  = 1 | s_j ) $ and the accuracy equals $s_j$ if $s_j \geq 0.5$ and $1-s_j$ otherwise. Therefore, in the perfectly calibrated case, we could empirically estimate accuracy  per group  for the unlabeled data using scores via $\hat{\theta}_g = (1/n_{U,g}) \sum_{j \in g}   s_j I(s_j \geq 0.5 ) + (1-s_j) I(s_j < 0.5)  $, where $n_{U,g}$ is the number of unlabeled examples that belong to group $g$. Metrics other than accuracy could also be estimated per group in a similar fashion.

In practice, however,  many classification models, particularly complex ones such as deep learning models, can be significantly miscalibrated (see, e.g.,~\cite{guo2017calibration, kull2017beta, kumar2019verified, ovadia2019}) and using the uncalibrated scores in such situations will lead to biased estimates of the true accuracy per group. The key idea of our approach is to use the labeled data to learn how to calibrate the scores such that the calibrated scores can contribute to less biased estimates of accuracy. Let $z_j = E[ I( \hat{y}_j = y_j ) ] = P(y_j = \hat{y}_j | s_j$ ) be the true (unknown) accuracy of the model given score $s_j$.
We treat each $z_j, j=1,\ldots,n_U$  as a latent variable per example. The high-level steps of the approach are as follows:
\begin{itemize}
\item We use the $n_L$ labeled examples to estimate groupwise calibration functions  with parameters $\phi_g$, that  transform the (potentially) uncalibrated scores $s$  of the model to calibrated scores. More specifically, we perform Bayesian inference (see Section \ref{sec:hierbayes} below) to obtain posterior samples from $P(\phi_g | D_L)$ for the groupwise calibration parameters   $\phi_g$. 
\item We then  obtain posterior samples from $P_{\phi_g}(z_j | D_L, s_j)$ for each unlabeled example $j=1,\ldots,n_U$,  conditioned on posterior samples of the $\phi_g$'s.
\item Finally, we use posterior samples from the $z_j$'s, combined with the labeled data, to generate  estimates of the groupwise metrics $\theta_g$ and the  difference in metrics $\Delta$. 
\end{itemize}

We can compute a posterior sample for ${\theta}^t_g$, given each set of posterior samples for $\phi_g^t$ and $z^t_1,\ldots,z^t_{n_U}$, by combining estimates of accuracies for the unlabeled examples with the observed outcomes for the labeled instances:
\begin{align}
 \theta^t_g = \frac{1}{n_{L,g} + n_{U,g}} \biggl( \sum_{i: i \in g}  I( \hat{y}_i = y_i ) +   \sum_{j:j \in g}  z^t_j \biggr)
\end{align}
where $t=1,...,T$ is an index over $T$ MCMC samples.  These posterior samples in turn can be used to generate an empirical posterior distribution $\{\Delta^1,\ldots,\Delta^T\}$ for $\Delta$, where $\Delta^t = \theta^t_1 - \theta^t_0$. Mean posterior estimates can be obtained by averaging over samples, i.e. $\hat{\Delta} = (1 / T) \sum^T_{t} \Delta^t$.  Even with very small amounts of labeled data (e.g., $n_L = 10)$  we will demonstrate later in the paper that we can make much more accurate inferences about fairness metrics via this Bayesian calibration approach, compared to using only the labeled data directly.

\subsection{Hierarchical Bayesian Calibration}
\label{sec:hierbayes}

\begin{figure}[t]
\centering 
   \subfloat{\includegraphics[width=\linewidth]{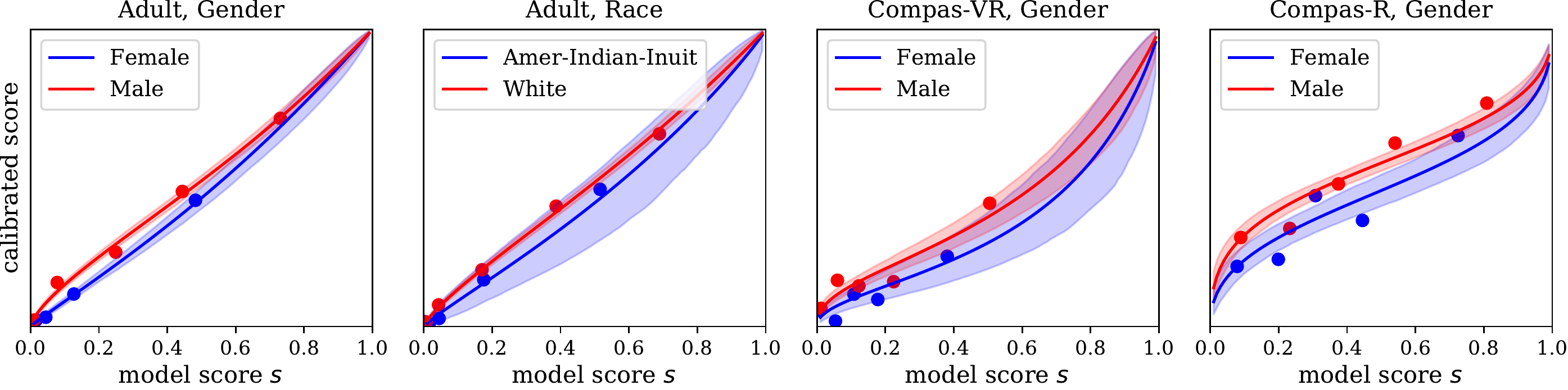}}
  \caption{Hierarchical Bayesian calibration of two demographic groups across four dataset-group pairs, with posterior means and 95\% credible intervals per group.  The $x$-axis is the model score $s$ for class $y=1$, and the $y$-axis is the calibrated score. Instances in each group are binned into 5 equal-sized bins by model score, and blue and red points show the fraction of positive samples per group for each bin.} 
  \label{fig:calibrationcurves}
\end{figure}

Bayesian calibration is a key step  in our approach above. We describe Bayesian inference below for the beta calibration model specifically~\citep{kull2017beta} but other calibration models could also be used. The beta calibration model  maps a score from a binary classifier with scores $s = P_M(y=1|x) \in [0,1]$  to a recalibrated score according to:
\begin{align}
f( s; a , b, c ) = \frac{ 1 }{ 1 + e^{ - c - a \log s + b \log ( 1- s ) }} 
\end{align}
 where $a$, $b$, and $c$ are calibration parameters with $a,b \geq 0$. 
 This model can capture  a wide variety of miscalibration patterns, producing the identity mapping if $s$ is already calibrated when $a=b=1, c=0$. Special cases of this model are the linear-log-odds (LLO) calibration model~\citep{turner2014forecast} when $a=b$, and temperature scaling~\citep{guo2017calibration} when $a=b, c=0$.

In our hierarchical Bayesian extension of the beta calibration model, we assume that each group (e.g., female, male)  is associated with its own set of calibration parameters $\phi_g = \{a_g,b_g,c_g\}$ and therefore each group can be miscalibrated in different ways (e.g., see Figure \ref{fig:calibrationcurves}). To apply this model to the observed data, we assume that the true labels for the observed instances are sampled according to:
\[
    y_i \sim \mathrm{Bernoulli}\bigl( f( s_i; a_{g_i},b_{g_i},c_{g_i} ) \bigr) 
 \]
where $g_i$ is the group associated with instance $i$, $1 \le i \le n_L$. 
For any unlabeled example $j = 1,\ldots, n_U$, conditioned on calibration parameters $\phi_{g_j}$ for the group for $j$,   we can compute the latent variable $z_j = f(s_j ; \ldots) I(s_j \ge 0.5) + (1 - f(s_j ; \ldots)) I(s_j < 0.5)$, i.e., the calibrated probability that the model's prediction on instance $j$ is correct.

We assume that the parameters from each individual group are sampled from a shared distribution:
$\log a_g \sim \mathrm{N}( \mu_a , \sigma_a ), \log b_g \sim \mathrm{N}( \mu_b , \sigma_b ), 
 c_g \sim \mathrm{N}( \mu_c , \sigma_c )$
where $\pi = \{\mu_a, \sigma_a, \mu_b, \sigma_b, \mu_c, \sigma_c \}$ is the set of hyperparameters of the shared distributions.  
We complete the hierarchical model by placing the following priors on the hyperparameters (TN is the truncated normal distribution):
\begin{align*} 
\mu_a \sim \mathrm{N}(0,.4),\mu_b \sim \mathrm{N}(0,.4),\mu_c \sim \mathrm{N}(0,2), 
\sigma_a \sim \mathrm{TN}(0,.15), \sigma_b \sim \mathrm{TN}(0,.15),\sigma_c \sim \mathrm{TN}(0,.75)
\end{align*}
These priors were chosen to place reasonable bounds on the calibration parameters and allow for diverse patterns of miscalibration (e.g., both over and under-confidence or a model) to be expressed a priori. 
We use exactly these same prior settings in all our experiments across all  datasets, all  groups, and all labeled and unlabeled dataset sizes, demonstrating the robustness of these settings across a wide variety of contexts. In addition,  the Supplement contains results of a sensitivity analysis for the variance parameters in the prior, illustrating robustness across a broad range of settings of these parameters.

The model was implemented as a graphical model (see Supplement) in JAGS, a common tool for Bayesian inference with Markov chain Monte Carlo~\citep{plummer2003jags}. All of the results in this paper are based on 4 chains, with 1500 burn-in iterations and 200 samples per chain, resulting in $T=800$ sample overall. These  hyperparameters were determined based on a few simulation runs across datasets, checking visually for lack of auto-correlation, with convergence assessed using the standard measure of within-to-between-chain variability. Although MCMC can sometimes be slow for high-dimensional problems,  with 100 labeled data points (for example) and  10k unlabeled data points the sampling procedure takes about 30 seconds (using non-optimized Python/JAGS code on a standard desktop computer) demonstrating the practicality of this procedure.

\paragraph{Theoretical Considerations:} 
Lemma~\ref{lemma:delta_bias} below relates potential error in the calibration mapping (e.g., due to misspecification of the parametric form of the calibration function $f(s;\ldots)$) to error in the estimate of $\Delta$ itself.

\begin{lemma} 
Given a prediction model $M$ and score distribution $P(s)$,
let $f_g(s;\phi_g): [0,1] \rightarrow [0,1]$ denote the calibration model for group $g$;
let $f^*_g(s):[0,1] \rightarrow [0,1]$  be the optimal calibration function which maps 
$s = P_{M}(\hat{y}=1|g)$ to $P(y=1|g)$;
and $\Delta^*$ is the true value of the metric.
Then the absolute error of the expected estimate w.r.t. $\phi$ can be bounded as:
$|\mathbb{E}_{\phi}\Delta - \Delta^*| \le \|\bar{f_0}-f^*_0\|_1 + \|\bar{f_1}-f^*_1\|_1$, 
where $\bar{f_g}(s) = \mathbb{E}_{\phi_g} f_g(s;{\phi_g}), \forall s\in[0,1]$, 
and $\|\cdot\|_1$ is the expected $L1$ distance  w.r.t. $P(s|g)$. (Proof provided in the Supplement).
\label{lemma:delta_bias}
\end{lemma}
Thus, reductions in the L1 calibration error directly reduce an upper bound on the L1 error in estimating $\Delta$. The results in  Figure~\ref{fig:calibrationcurves} suggest that even with the relatively simple parametric beta calibration method, the error in calibration (difference between the fitted calibration functions) (blue and red curves) and the empirical data (blue and red dots) is quite low across all 4 datasets. The possibility of using more flexible calibration functions is an  interesting direction for future work.

\begin{table}[t!]
\caption{Datasets used in the paper. $G$ is the sensitive attribute, $P(g=0)$ is the probability of the privileged group, and $P(y=1)$ is the probability of the positive label for classification. The privileged groups $g=0$ are gender: male,
age: senior or adult, and race: white or Caucasian.}
\centering
\resizebox{0.65\textwidth}{!}{  
\begin{tabular}{cccccc}
\toprule
   Dataset &  Test Size &           $G$ &    $P(g=0)$ &  $P(y=1)$ \\
\midrule
     Adult &     10054 &  gender, race &    0.68, 0.86 &      0.25 \\
      Bank &     13730 &           age &          0.45 &      0.11 \\
    German &       334 &   age, gender &    0.79, 0.37 &      0.17 \\
  Compas-R &      2056 &  gender, race &     0.7, 0.85 &      0.69 \\
 Compas-VR &      1337 &  gender, race &     0.8, 0.34 &      0.47 \\
     Ricci &        40 &          race &          0.65 &      0.50 \\
\bottomrule
\end{tabular}}
\label{tab:data}
\end{table}

\section{Datasets, Classification Models, and Illustrative Results}
\label{sec:illustration}

One of the main goals of our experiments is to assess the  accuracy of different estimation  methods, using relatively limited amounts of labeled data, relative to the true value of the metric. By ``true value" we mean the value we could measure on an infinitely large test sample. Since such a sample is not available, we use as a proxy the value of metric computed on all of the test set for each dataset in our experiments.

We followed the experimental methods and used the code for preprocessing and training prediction models from~\cite{friedler2019comparative} who systematically compared fairness metrics and fairness-aware algorithms across a variety of datasets.
Specifically, we use the 
Adult, 
German Credit, 
Ricci, 
and Compas datasets (for recidivism and violent recidivism), 
all used in~\cite{friedler2019comparative}, 
as well as the Bank Telemarketing dataset~\citep{moro2014data}.
Summary statistics for the datasets are shown in Table \ref{tab:data}. Classification models (logistic regression, multilayer perceptron (MLP) with a single hidden layer, random forests, Gaussian Naive Bayes) were trained using standard default parameter settings with the code provided by~\cite{friedler2019comparative} and predictions generated on the test data. Sensitive attributes are not included as inputs to the models. Unless a specific train/test split is provided in the original dataset, we randomly sample 2/3 of the instances for training and 1/3 for test.  
Additional details on models and datasets are provided in the Supplement.

\begin{figure}[t]
\centering
  \includegraphics[width=\linewidth]{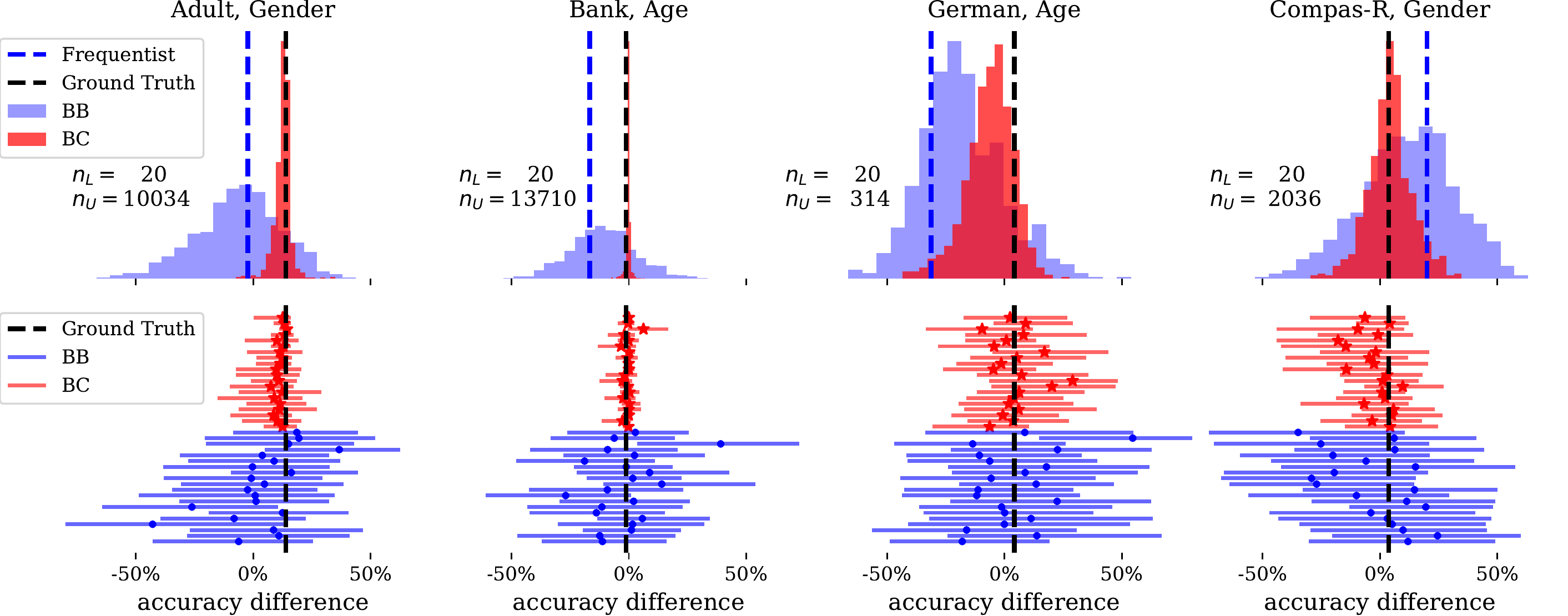}
  \caption{Posterior density (samples) and frequentist estimates (dotted vertical blue lines) for the difference in group accuracy $\Delta$ for 4 datasets with $n_L = 20$ random labeled examples   for both the BB (beta-binomial) and BC (Bayesian calibration) methods. Ground truth is a vertical black line. The underlying model is an MLP. The 20 examples were randomly sampled 20 different times.
 Upper plots show the histograms of posterior samples for the first sample, lower plots show the 95\% posterior credible intervals for all 20 runs, where the x-axis is $\Delta$.}
  \label{fig:simulation} 
\end{figure}

\paragraph{Illustrative Results:} To illustrate our approach we compare the results of the frequentist,  beta-binomial (BB), and Bayesian calibration (BC) approaches for assessing group  differences in accuracy across 4 datasets, for a multilayer perceptron (MLP) binary classifier. 
We ran the methods on 20 runs of randomly sampled  sets of $n_L = 20$ labeled examples. The BC method was given access to the remaining $n_U$ unlabeled test examples minus the 20 labeled examples for each run, as described in Table \ref{tab:data}. We define ground truth as the frequentist $\Delta$ value computed on all the labeled data in the test set.  
Figure \ref{fig:simulation} shows the results across the 4 datasets. The top figure corresponds to the first run out of 20 runs, showing the histogram of 800 posterior samples from the BB (blue) and BC (red) methods. 
The lower row of plots summarizes the results for all 20 runs, showing the 95\% posterior credible intervals (CIs) (red and blue horizontal lines for BC and BB respectively) along with posterior means (red and blue marks). 

Because of the relatively weak prior (Beta(1,1) on group accuracy) the posterior means of the BB samples tend to be relatively close to the frequentist estimate (light and dark blue respectively) on each run and both can be relatively far away from ground truth value for $\Delta$ (in black). Although the BB method is an improvement over being frequentist, in that it provides posterior uncertainty about $\Delta$, it nonetheless has high variance (locations of the posterior means) as well as high posterior uncertainty (relatively wide CIs). The BC method in contrast, by using the unlabeled data in addition to the labeled data,  produces  posterior estimates where the mean tends to be much closer to ground truth than BC.

The posterior information about $\Delta$ can be used to provide users with a summary report
that includes information about the direction of potential bias (e.g., $P(\Delta>0| D_L, D_U)$, the degree of bias (e.g., via the MPE $\hat{\Delta}$), 95\% posterior CIs on  $\Delta$, and the probability that the model is ``practically fair" (assessed via $P(|\Delta|<\epsilon| D_L, D_U)$, e.g., see~\cite{benavoli2017time}).
For example with BC, given the observed data, practitioners can conclude from the information in the upper row of Figure~\ref{fig:simulation}, and with $\epsilon=0.02$, that there is a 0.99 probability for the Adult data  that the classifier is more accurate for females than males; and with probability 0.87 that the classifier is practically fair with respect to accuracy for junior and senior individuals in the Bank data.

\section{Experiments and Results}
 
\begin{figure}[!t]
\centering
  \includegraphics[width=\linewidth]{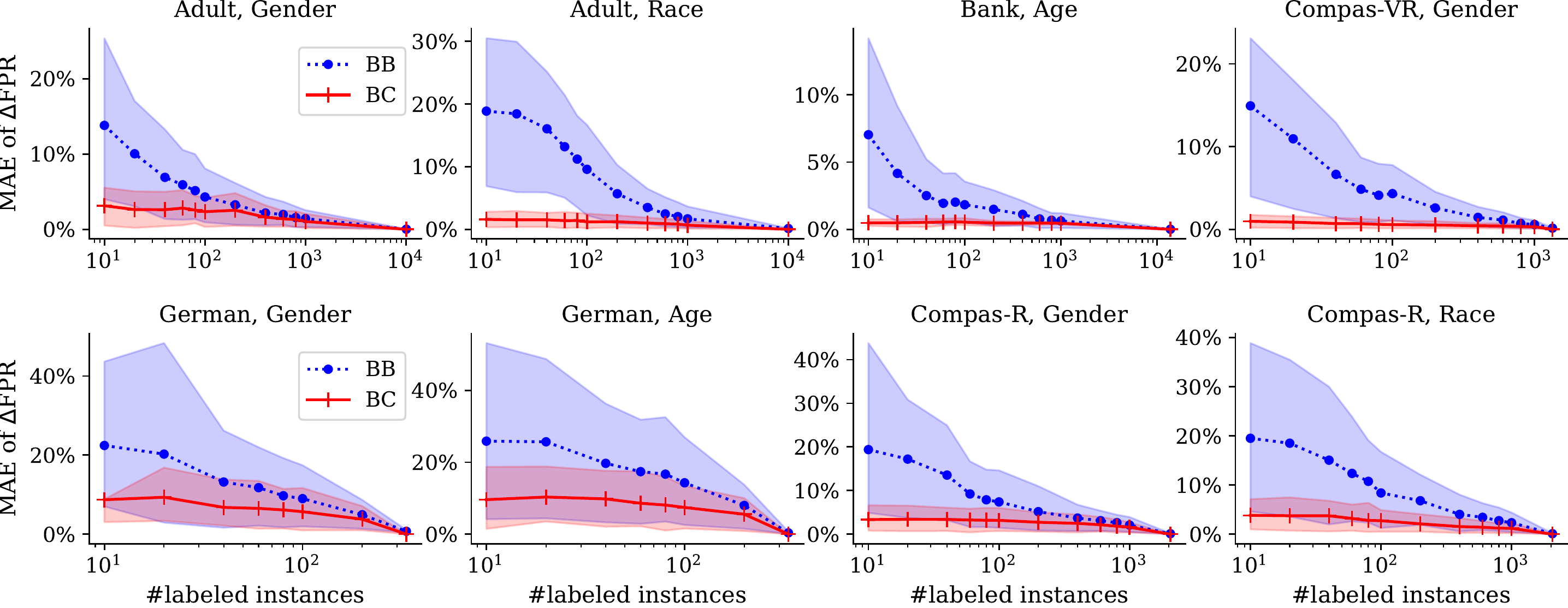}
   \caption{Mean absolute error (MAE) of the difference between algorithm estimates and ground truth for group difference in FPR, as a function of number of labeled instances, for 8 different dataset-group pairs.  Shading indicates 95\% error bars for each method. }
  \label{fig:MAEgraph}
\end{figure}

In this section we systematically evaluate the quality of different estimation approaches by repeating the same type of experiment as in Section \ref{sec:illustration} and Figure \ref{fig:simulation} across different amounts of labeled data $n_L$.
In particular, for each value of $n_L$ we randomly sample sets of labeled datasets of size $n_L$, generate point estimates of a metric $\Delta$ of interest for each labeled dataset for each of the BB and BC estimation methods, and compute the mean absolute error (MAE) between the point estimates and the true value (computed on the full labeled test set). The frequency-based estimates are not shown for clarity---they are almost always worse than both BB and BC. As an example, Figure \ref{fig:MAEgraph} illustrates the quality of estimation where $\Delta$ is the FPR group difference $\Delta$  for the MLP classification model, evaluated across 8 different dataset-group pairs. Each y-value is the average of 100 different randomly sampled sets of $n_L$ instances, where $n_L$ is the corresponding x-axis value. 
The BC method dominates BB   across all datasets indicating that the calibrated scores are very effective at improving the accuracy in estimating group FPR. This is particularly true for small amounts of labeled data (e.g., up to $n_L = 100$) where the BB Method can be highly inaccurate, e.g., MAEs on the order of 10 or 20\% when the true value of $\Delta$ is often less than 10\%.

\begin{table}[!h]
\caption{{\bf MAE for $\Delta$ Accuracy Estimates},   with $n_L=10$, across 100 runs of labeled samples, for 4 different trained models (groups of columns) and 10 different dataset-group combinations (rows). Lowest error rate per row-col group in bold if the difference among methods are statistically significant under Wilcoxon signed-rank test (p=0.05). Estimation methods are Freq (Frequentist), BB, and BC. Freq and BB use only labeled samples, BC uses both labeled samples and unlabeled data.
Trained models are Multilayer Perceptron, Logistic Regression, Random Forests, and Gaussian Naive Bayes.
}
\resizebox{\textwidth}{!}{
\begin{tabular}{@{}rrrrrcrrrcrrrcrrc@{}}
\toprule 
& \phantom{a} &  \multicolumn{3}{c}{Multi-layer Perceptron}  
& \phantom{a}&  \multicolumn{3}{c}{Logistic Regression}
& \phantom{a} & \multicolumn{3}{c}{Random Forest} 
& \phantom{a} & \multicolumn{3}{c}{Gaussian Naive Bayes}\\ 
\cmidrule{3-5} \cmidrule{7-9} \cmidrule{11-13} \cmidrule{15-17}
Dataset, Attribute && Freq &BB  &BC && Freq &BB  &BC && Freq &BB  &BC && Freq &BB  &BC \\ \midrule
Adult, Race 		&&16.5 &18.5 &\textbf{ 3.9}		&&16.4 &18.7 &\textbf{ 2.9}		&&16.5 &18.2 &\textbf{ 3.2}		&&17.6 &18.9 &\textbf{ 3.6}\\ 
Adult, Gender 		&&19.7 &17.4 &\textbf{ 5.1}		&&19.1 &16.1 &\textbf{ 2.2}		&&17.7 &17.4 &\textbf{ 4.8}		&&19.7 &16.2 &\textbf{ 5.4}\\ 
 Bank, Age 		&&15.9 &13.9 &\textbf{ 2.5}		&&13.9 &13.0 &\textbf{ 1.4}		&&11.8 &11.1 &\textbf{ 1.0}		&&15.5 &13.7 &\textbf{ 1.7}\\ 
German, Age 		&&34.6 &19.8 &\textbf{ 5.0}		&&37.1 &21.2 &\textbf{ 8.7}		&&33.6 &18.7 &\textbf{ 8.2}		&&36.6 &20.4 &\textbf{11.5}\\ 
German, Gender 		&&30.7 &21.6 &\textbf{ 8.2}		&&25.6 &17.4 &\textbf{ 6.3}		&&27.7 &19.3 &\textbf{ 8.6}		&&30.0 &20.1 &\textbf{ 6.5}\\ 
Compas-R, Race 		&&31.5 &21.0 &\textbf{ 4.2}		&&31.7 &20.4 &\textbf{ 4.8}		&&29.3 &20.3 &\textbf{ 2.4}		&&33.5 &23.2 &\textbf{ 8.4}\\ 
Compas-R, Gender 		&&33.7 &21.6 &\textbf{ 5.0}		&&34.3 &21.9 &\textbf{ 3.8}		&&36.3 &23.3 &\textbf{ 4.4}		&&40.5 &25.5 &\textbf{13.7}\\ 
Compas-VR, Race 		&&18.7 &17.1 &\textbf{ 4.0}		&&18.5 &15.6 &\textbf{ 4.4}		&&18.2 &15.8 &\textbf{ 2.4}		&&26.6 &19.8 &\textbf{ 6.5}\\ 
Compas-VR, Gender 		&&20.6 &16.9 &\textbf{ 5.4}		&&19.9 &16.6 &\textbf{ 5.3}		&&22.3 &19.0 &\textbf{ 6.3}		&&31.3 &21.5 &\textbf{ 9.8}\\ 
Ricci, Race 		&&23.5 &17.7 &\textbf{14.6}		&&14.6 &14.6 &\textbf{ 7.9}		&& 6.3 &12.2 &\textbf{ 2.1}		&& 8.9 &13.1 &\textbf{ 1.6}\\ 
\bottomrule
\end{tabular}}
\label{tab:acc}

\caption{{\bf MAE for $\Delta$ TPR Estimates}, with $n_L=200$. Same setup as for Table \ref{tab:acc}. Compas-VR race and Ricci race are not included since there are no positive instances for some groups, and some entries under Freq cannot be estimated for the same reason.}
\resizebox{\textwidth}{!}{
\begin{tabular}{@{}rrrrrcrrrcrrrcrrc@{}}
\toprule 
& \phantom{a} &  \multicolumn{3}{c}{Multi-layer Perceptron}  
& \phantom{a}&  \multicolumn{3}{c}{Logistic Regression}
& \phantom{a} & \multicolumn{3}{c}{Random Forest} 
& \phantom{a} & \multicolumn{3}{c}{Gaussian Naive Bayes}\\ 
\cmidrule{3-5} \cmidrule{7-9} \cmidrule{11-13} \cmidrule{15-17}
Dataset, Attribute && Freq &BB  &BC && Freq &BB  &BC && Freq &BB  &BC && Freq &BB  &BC \\ \midrule
Adult, Race 		&&\textemdash&12.5 &\textbf{ 5.8}		&&\textemdash&14.7 &\textbf{ 7.0}		&&\textemdash&14.3 &\textbf{ 4.6}		&&\textemdash&14.6 &\textbf{ 3.0}\\ 
Adult, Gender 		&&16.3 &14.3 &\textbf{ 4.3}		&&15.8 &14.0 &\textbf{ 4.6}		&&16.1 &14.2 &\textbf{ 7.3}		&&15.0 &13.4 &11.5\\ 
 Bank, Age 		&&16.8 &15.0 &\textbf{ 4.8}		&&17.7 &15.9 &\textbf{ 4.2}		&&16.6 &14.9 &\textbf{ 3.1}		&&17.3 &15.7 &\textbf{ 2.3}\\ 
German, Age 		&& 4.7 & 4.7 &\textbf{ 3.0}		&& 5.6 & 5.4 &\textbf{ 2.6}		&& 5.1 & 5.1 &\textbf{ 3.1}		&& 6.8 & 6.5 &\textbf{ 2.8}\\ 
German, Gender 		&& \textbf{0.7} & 1.0 &1.6		&& 3.3 & 3.3 &\textbf{ 2.1}		&& 3.1 & 3.2 &\textbf{ 2.1}		&& 4.8 & 4.7 &\textbf{ 2.2}\\ 
Compas-R, Race 		&&\textemdash& 7.6 &\textbf{ 2.5}		&&\textemdash& 7.9 &\textbf{ 2.6}		&&\textemdash& 9.2 &\textbf{ 2.1}		&&\textemdash& 4.5 &\textbf{ 2.0}\\ 
Compas-R, Gender 		&&10.0 & 9.5 &\textbf{ 1.9}		&&10.0 & 9.4 &\textbf{ 1.8}		&&11.3 &10.7 &\textbf{ 2.6}		&& 5.6 & 5.5 &\textbf{ 0.3}\\ 
Compas-VR, Gender 		&&14.9 &12.2 &\textbf{ 2.9}		&& 8.9 &10.7 &\textbf{ 2.0}		&&14.6 &10.5 &7.2		&&12.5 &10.0 &\textbf{ 1.3}\\ 
\bottomrule
\end{tabular}
}
\label{tab:tpr}
\end{table}

In the Supplement we show that the trend of results shown in Figure \ref{fig:MAEgraph}, namely that BC produces significantly more accurate estimates of group fairness metrics $\Delta$, is replicated across all 4 classification models that we investigated, across FPR, TPR and Accuracy metrics, and across all datasets. To summarize the full set of results we show a subset in tabular form, across all 4 classification models and 10 dataset-group pairs, with $n_L$ fixed:  Table \ref{tab:acc} for Accuracy with $n_L=10$ and  Table \ref{tab:tpr} for TPR with $n_L=200$. (We used larger $n_L$ values for TPR and FPR than for accuracy in the results above since TPR and FPR depend on estimating conditional probabilities that can have zero supporting counts in the labeled data, causing a problem for frequentist estimators). The results above and in the Supplement demonstrate the significant gains in accuracy that can be achieved with the proposed approach.  We also evaluated the effect of using LLO calibration instead of beta calibration methods and found little difference between the two methods (details in Supplement). 

For concreteness we demonstrated our results with three popular fairness metrics ($\Delta$ accuracy, TPR, and FPR) in the paper. However, we can directly extend this approach to handle metrics such as calibration and balance~\citep{kleinberg2016inherent} as well as ratio-based metrics. In particular, by predicting the distribution of class labels $y$ with the calibrated model scores, any fairness metric that can be   defined as a deterministic function of calibrated model scores $s$, labels $y$ and groups $g$ can levarage unlabeled data to reduce variance using our proposed method.

Consideration of the bias-variance properties of the different methods reveals a fundamental tradeoff. The labeled data contribute no bias to the estimate but can have high variance for small $n_L$, whereas the unlabeled data (via their calibrated scores) contribute little variance but can have a persistent bias due to potential misspecification in the parametric calibration model. An open question, that is beyond the scope of this paper, is how to balance this bias-variance tradeoff in a more adaptive fashion as a function of $n_L$ and $n_U$, to further improve the accuracy of estimates of fairness metrics for arbitrary datasets. One potential option would be to a more flexible calibration method (e.g., Gaussian process calibration as proposed in~\cite{wenger2020non}). Another option would be to automatically quantify the calibration bias and tradeoff the contributions of labeled and unlabeled data accordingly in estimating $\theta_g$'s and $\Delta$.

We also found empirically that while the posterior credible intervals (CIs) for the BB method are well-calibrated, those for BC tended to be overconfident as $n_L$ increases (see Supplement for details). This is likely  due to misspecification in the parametric beta calibration model. 
An interesting and important direction for future work is to develop methods that are better calibrated in terms of posterior credible intervals (e.g.,~\cite{syring2019calibrating})  and that can retain the low-variance advantages of the BC approach we propose here.

\section{Related Work}
 
Our Bayesian calibration approach builds on the work of~\cite{turner2014forecast} who used  hierarchical Bayesian methods for calibration of human judgement data using the LLO calibration model.  Bayesian approaches to classifier calibration include marginalizing over binned model scores~\citep{naeini2015obtaining} and calibration based on Gaussian processes~\citep{wenger2020non}. The Bayesian framework of~\cite{welinder2013lazy} in particular is close  in spirit to our work in that unlabeled examples are used to improve calibration,  but differs in that a generative mixture model is used for modeling of scores rather than direct calibration.  None of this prior work on Bayesian calibration addresses  fairness assessment and none (apart from~\cite{welinder2013lazy}) leverages unlabeled data.

There has also been work on uncertainty-aware assessment of classifier performance such as the use of Bayesian inference for classifier-related metrics such as  marginal accuracy~\citep{benavoli2017time} and precision-recall~\citep{goutte2005probabilistic}. Although these approaches share similarities with our work, they do not make use of unlabeled data. In contrast, the Bayesian evaluation methods proposed by~\cite{johnson2019gold} can use unlabeled data but   makes  strong prior assumptions that are specific to the application domain of diagnostic testing.  More broadly, other general approaches have been proposed for label-efficient classifier assessment including stratified sampling~\citep{sawade2010}, importance sampling~\citep{kumar2018classifier}, and active assessment with Thompson sampling~\citep{ji2020active}. 
All of these ideas could in principle be used in conjunction with our approach to further reduce estimation error. 

In the literature on algorithmic fairness there has been little prior work on uncertainty-aware assessment of fairness metrics---one exception is the proposed use of frequentist confidence interval methods for groupwise fairness in~\cite{besse2018confidence}. 
~\cite{dimitrakakis2019bayesian} proposed a framework called ``Bayesian fairness," but focused on decision-theoretic aspects of the problem rather than estimation of metrics.  ~\cite{foulds2020bayesian} developed Bayesian approches for  for parametric smoothing across groups to improve the quality of estimation of intersectional fairness metrics.
However, none of this work makes use of unlabeled data to improve  fairness assessment. And while there is prior work in fairness on leveraging unlabeled data~\citep{chzhen2019leveraging,noroozi2019leveraging,wick2019unlocking,zhang2020fairness}, the goal of that work has been  to produce   classifiers that are fair, rather than to assess the fairness of existing classifiers.

Finally, there is   recent concurrent work from a frequentist perspective  that uses Bernstein inequalities and knowledge of group proportions to upper bound the probability that the difference between the frequentist estimate of $\Delta$ and the true $\Delta$ exceeds some value~\citep{ethayarajh2020your}. While this work differs from our approach in that it does not explore the use of unlabeled data, the same broad conclusion is reached, namely that there can be high uncertainty in empirical estimates of groupwise fairness metrics, given the typical sizes of datasets used in machine learning.

\section{Conclusions} 
To answer to the question ``can I trust my fairness metric," we have stressed the importance of being aware of uncertainty in fairness assessment, especially when test sizes are relatively small (as is often the case in practice).  
To address this issue we propose a framework for combining labeled and unlabeled data to reduce estimation variance, using  Bayesian calibration of model scores on unlabeled data. The results demonstrate that the proposed method can systematically produce significantly more accurate estimates of fairness metrics, when compared to only using labeled data, across multiple different classification models, datasets, and sensitive attributes. The framework is straightforward to apply in practice and easy to extend to problems such as intersectional fairness (where estimation uncertainty is likely a significant issue) and to evaluation of fairness-aware algorithms. 

\section*{Acknowledgements}
This material is based upon work supported in part by the National Science Foundation under grants number 1900644 and 1927245 and by a Qualcomm Faculty Award (PS). In addition this work was partially funded by the Center for Statistics and Applications in Forensic Evidence (CSAFE) through Cooperative Agreement 70NANB20H019 between NIST and Iowa State University, which includes activities carried out at the University of California,  Irvine.

\newpage
\bibliography{fairness}

\begin{thebibliography}{41}
\providecommand{\natexlab}[1]{#1}
\providecommand{\url}[1]{\texttt{#1}}
\expandafter\ifx\csname urlstyle\endcsname\relax
  \providecommand{\doi}[1]{doi: #1}\else
  \providecommand{\doi}{doi: \begingroup \urlstyle{rm}\Url}\fi

\bibitem[Angwin et~al.(2017)Angwin, Larson, Mattu, and Kirchner]{angwin2017we}
Julia Angwin, Jeff Larson, Surya Mattu, and Lauren Kirchner.
\newblock How we analyzed the {COMPAS} recidivism algorithm.
\newblock \emph{URL https://www. propublica.
  org/article/how-we-analyzed-the-compas-recidivism-algorithm}, 2017.

\bibitem[Benavoli et~al.(2017)Benavoli, Corani, Dem{\v{s}}ar, and
  Zaffalon]{benavoli2017time}
Alessio Benavoli, Giorgio Corani, Janez Dem{\v{s}}ar, and Marco Zaffalon.
\newblock Time for a change: a tutorial for comparing multiple classifiers
  through {B}ayesian analysis.
\newblock \emph{The Journal of Machine Learning Research}, 18\penalty0
  (1):\penalty0 2653--2688, 2017.

\bibitem[Berk et~al.(2018)Berk, Heidari, Jabbari, Kearns, and
  Roth]{berk2018fairness}
Richard Berk, Hoda Heidari, Shahin Jabbari, Michael Kearns, and Aaron Roth.
\newblock Fairness in criminal justice risk assessments: The state of the art.
\newblock \emph{Sociological Methods \& Research}, 2018.

\bibitem[Besse et~al.(2018)Besse, del Barrio, Gordaliza, and
  Loubes]{besse2018confidence}
Philippe Besse, Eustasio del Barrio, Paula Gordaliza, and Jean-Michel Loubes.
\newblock Confidence intervals for testing disparate impact in fair learning.
\newblock \emph{arXiv preprint arXiv:1807.06362}, 2018.

\bibitem[Beutel et~al.(2019)Beutel, Chen, Doshi, Qian, Woodruff, Luu,
  Kreitmann, Bischof, and Chi]{beutel2019putting}
Alex Beutel, Jilin Chen, Tulsee Doshi, Hai Qian, Allison Woodruff, Christine
  Luu, Pierre Kreitmann, Jonathan Bischof, and Ed~H Chi.
\newblock Putting fairness principles into practice: Challenges, metrics, and
  improvements.
\newblock In \emph{Proceedings of the AAAI/ACM Conference on AI, Ethics, and
  Society}, pages 453--459, 2019.

\bibitem[Calders and Verwer(2010)]{calders2010three}
Toon Calders and Sicco Verwer.
\newblock Three naive {B}ayes approaches for discrimination-free
  classification.
\newblock \emph{Data Mining and Knowledge Discovery}, 21\penalty0 (2):\penalty0
  277--292, 2010.

\bibitem[Chen et~al.(2019)Chen, Szolovits, and Ghassemi]{chen2019can}
Irene~Y Chen, Peter Szolovits, and Marzyeh Ghassemi.
\newblock Can {AI} help reduce disparities in general medical and mental health
  care?
\newblock \emph{AMA Journal of Ethics}, 21\penalty0 (2):\penalty0 167--179,
  2019.

\bibitem[Chouldechova(2017)]{chouldechova2017fair}
Alexandra Chouldechova.
\newblock Fair prediction with disparate impact: A study of bias in recidivism
  prediction instruments.
\newblock \emph{Big Data}, 5\penalty0 (2):\penalty0 153--163, 2017.

\bibitem[Chzhen et~al.(2019)Chzhen, Denis, Hebiri, Oneto, and
  Pontil]{chzhen2019leveraging}
Evgenii Chzhen, Christophe Denis, Mohamed Hebiri, Luca Oneto, and Massimiliano
  Pontil.
\newblock Leveraging labeled and unlabeled data for consistent fair binary
  classification.
\newblock In \emph{Advances in Neural Information Processing Systems}, pages
  12739--12750, 2019.

\bibitem[Corbett-Davies and Goel(2018)]{corbett2018measure}
Sam Corbett-Davies and Sharad Goel.
\newblock The measure and mismeasure of fairness: A critical review of fair
  machine learning.
\newblock \emph{arXiv preprint arXiv:1808.00023}, 2018.

\bibitem[Dimitrakakis et~al.(2019)Dimitrakakis, Liu, Parkes, and
  Radanovic]{dimitrakakis2019bayesian}
Christos Dimitrakakis, Yang Liu, David~C Parkes, and Goran Radanovic.
\newblock Bayesian fairness.
\newblock In \emph{Proceedings of the AAAI Conference on Artificial
  Intelligence}, volume~33, pages 509--516, 2019.

\bibitem[Dwork et~al.(2012)Dwork, Hardt, Pitassi, Reingold, and
  Zemel]{dwork2012fairness}
Cynthia Dwork, Moritz Hardt, Toniann Pitassi, Omer Reingold, and Richard Zemel.
\newblock Fairness through awareness.
\newblock In \emph{Proceedings of the 3rd Innovations in Theoretical Computer
  Science Conference}, pages 214--226, 2012.

\bibitem[Ethayarajh(2020)]{ethayarajh2020your}
Kawin Ethayarajh.
\newblock Is your classifier actually biased? {M}easuring fairness under
  uncertainty with bernstein bounds.
\newblock In \emph{Proceedings of the Annual Meeting of the Association for
  Computational Linguistics}, pages 2914--2919, 2020.

\bibitem[Feldman et~al.(2015)Feldman, Friedler, Moeller, Scheidegger, and
  Venkatasubramanian]{feldman2015certifying}
Michael Feldman, Sorelle~A Friedler, John Moeller, Carlos Scheidegger, and
  Suresh Venkatasubramanian.
\newblock Certifying and removing disparate impact.
\newblock In \emph{Proceedings of the 21th ACM SIGKDD International Conference
  on Knowledge Discovery and Data Mining}, pages 259--268, 2015.

\bibitem[Foulds et~al.(2020)Foulds, Islam, Keya, and Pan]{foulds2020bayesian}
James~R Foulds, Rashidul Islam, Kamrun~Naher Keya, and Shimei Pan.
\newblock Bayesian modeling of intersectional fairness: The variance of bias.
\newblock In \emph{Proceedings of the SIAM International Conference on Data
  Mining}, pages 424--432, 2020.

\bibitem[Friedler et~al.(2019)Friedler, Scheidegger, Venkatasubramanian,
  Choudhary, Hamilton, and Roth]{friedler2019comparative}
Sorelle~A Friedler, Carlos Scheidegger, Suresh Venkatasubramanian, Sonam
  Choudhary, Evan~P Hamilton, and Derek Roth.
\newblock A comparative study of fairness-enhancing interventions in machine
  learning.
\newblock In \emph{Proceedings of the Conference on Fairness, Accountability,
  and Transparency}, pages 329--338, 2019.

\bibitem[Gelman et~al.(2013)Gelman, Carlin, Stern, Dunson, Vehtari, and
  Rubin]{gelman2013bayesian}
Andrew Gelman, John~B Carlin, Hal~S Stern, David~B Dunson, Aki Vehtari, and
  Donald~B Rubin.
\newblock \emph{Bayesian Data Analysis}.
\newblock CRC press, 2013.

\bibitem[Goutte and Gaussier(2005)]{goutte2005probabilistic}
Cyril Goutte and Eric Gaussier.
\newblock A probabilistic interpretation of precision, recall and {F}-score,
  with implication for evaluation.
\newblock In \emph{Proceedings of the European Conference on Information
  Retrieval}, pages 345--359, 2005.

\bibitem[Guo et~al.(2017)Guo, Pleiss, Sun, and Weinberger]{guo2017calibration}
Chuan Guo, Geoff Pleiss, Yu~Sun, and Kilian~Q Weinberger.
\newblock On calibration of modern neural networks.
\newblock In \emph{Proceedings of the 34th International Conference on Machine
  Learning}, pages 1321--1330, 2017.

\bibitem[Hardt et~al.(2016)Hardt, Price, and Srebro]{hardt2016equality}
Moritz Hardt, Eric Price, and Nati Srebro.
\newblock Equality of opportunity in supervised learning.
\newblock In \emph{Advances in Neural Information Processing Systems}, pages
  3315--3323, 2016.

\bibitem[Ji et~al.(2020)Ji, Logan~IV, Smyth, and Steyvers]{ji2020active}
Disi Ji, Robert~L Logan~IV, Padhraic Smyth, and Mark Steyvers.
\newblock Active {B}ayesian assessment for black-box classifiers.
\newblock \emph{arXiv preprint arXiv:2002.06532}, 2020.

\bibitem[Johnson et~al.(2019)Johnson, Jones, and Gardner]{johnson2019gold}
Wesley~O Johnson, Geoff Jones, and Ian~A Gardner.
\newblock Gold standards are out and {B}ayes is in: Implementing the cure for
  imperfect reference tests in diagnostic accuracy studies.
\newblock \emph{Preventive Veterinary Medicine}, 167:\penalty0 113--127, 2019.

\bibitem[Kamishima et~al.(2012)Kamishima, Akaho, Asoh, and
  Sakuma]{kamishima2012fairness}
Toshihiro Kamishima, Shotaro Akaho, Hideki Asoh, and Jun Sakuma.
\newblock Fairness-aware classifier with prejudice remover regularizer.
\newblock In \emph{Proceedings of the Joint European Conference on Machine
  Learning and Knowledge Discovery in Databases}, pages 35--50. Springer, 2012.

\bibitem[Kleinberg et~al.(2016)Kleinberg, Mullainathan, and
  Raghavan]{kleinberg2016inherent}
Jon Kleinberg, Sendhil Mullainathan, and Manish Raghavan.
\newblock Inherent trade-offs in the fair determination of risk scores.
\newblock \emph{arXiv preprint arXiv:1609.05807}, 2016.

\bibitem[Kull et~al.(2017)Kull, Silva~Filho, and Flach]{kull2017beta}
Meelis Kull, Telmo Silva~Filho, and Peter Flach.
\newblock Beta calibration: a well-founded and easily implemented improvement
  on logistic calibration for binary classifiers.
\newblock In \emph{Proceedings of the International Conference on Artificial
  Intelligence and Statistics}, pages 623--631, 2017.

\bibitem[Kumar et~al.(2019)Kumar, Liang, and Ma]{kumar2019verified}
Ananya Kumar, Percy~S Liang, and Tengyu Ma.
\newblock Verified uncertainty calibration.
\newblock In \emph{Advances in Neural Information Processing Systems}, pages
  3787--3798, 2019.

\bibitem[Kumar and Raj(2018)]{kumar2018classifier}
Anurag Kumar and Bhiksha Raj.
\newblock Classifier risk estimation under limited labeling resources.
\newblock In \emph{Proceedings of the Pacific-Asia Conference on Knowledge
  Discovery and Data Mining}, pages 3--15. Springer, 2018.

\bibitem[Moro et~al.(2014)Moro, Cortez, and Rita]{moro2014data}
S{\'e}rgio Moro, Paulo Cortez, and Paulo Rita.
\newblock A data-driven approach to predict the success of bank telemarketing.
\newblock \emph{Decision Support Systems}, 62:\penalty0 22--31, 2014.

\bibitem[Naeini et~al.(2015)Naeini, Cooper, and
  Hauskrecht]{naeini2015obtaining}
Mahdi~Pakdaman Naeini, Gregory~F Cooper, and Milos Hauskrecht.
\newblock Obtaining well calibrated probabilities using {B}ayesian binning.
\newblock In \emph{Proceedings of the AAAI Conference}, pages 2901--2907, 2015.

\bibitem[Noroozi et~al.(2019)Noroozi, Bahaadini, Sheikhi, Mojab, and
  Philip]{noroozi2019leveraging}
Vahid Noroozi, Sara Bahaadini, Samira Sheikhi, Nooshin Mojab, and S~Yu Philip.
\newblock Leveraging semi-supervised learning for fairness using neural
  networks.
\newblock In \emph{Proceedings of the IEEE International Conference On Machine
  Learning And Applications}, pages 50--55, 2019.

\bibitem[Ovadia et~al.(2019)Ovadia, Fertig, Lakshminarayanan, Nowozin, Sculley,
  Dillon, Ren, Nado, and Snoek]{ovadia2019}
Yaniv Ovadia, Emily Fertig, Balaji Lakshminarayanan, Sebastian Nowozin,
  D~Sculley, Joshua Dillon, Jie Ren, Zachary Nado, and Jasper Snoek.
\newblock Can you trust your model's uncertainty? {E}valuating predictive
  uncertainty under dataset shift.
\newblock In \emph{Advances in Neural Information Processing Systems}, pages
  13969--13980, 2019.

\bibitem[Plummer(2003)]{plummer2003jags}
Martyn Plummer.
\newblock {JAGS}: A program for analysis of {B}ayesian graphical models using
  {G}ibbs sampling.
\newblock In \emph{Proceedings of the International Workshop on Distributed
  Statistical Computing}, 2003.

\bibitem[Rajkomar et~al.(2018)Rajkomar, Hardt, Howell, Corrado, and
  Chin]{rajkomar2018ensuring}
Alvin Rajkomar, Michaela Hardt, Michael~D Howell, Greg Corrado, and Marshall~H
  Chin.
\newblock Ensuring fairness in machine learning to advance health equity.
\newblock \emph{Annals of Internal Medicine}, 169\penalty0 (12):\penalty0
  866--872, 2018.

\bibitem[Sawade et~al.(2010)Sawade, Landwehr, Bickel, and Scheffer]{sawade2010}
Christoph Sawade, Niels Landwehr, Steffen Bickel, and Tobias Scheffer.
\newblock Active risk estimation.
\newblock In \emph{Proceedings of the 27th International Conference on Machine
  Learning}, pages 951--958, 2010.

\bibitem[Syring and Martin(2019)]{syring2019calibrating}
Nicholas Syring and Ryan Martin.
\newblock Calibrating general posterior credible regions.
\newblock \emph{Biometrika}, 106\penalty0 (2):\penalty0 479--486, 2019.

\bibitem[Turner et~al.(2014)Turner, Steyvers, Merkle, Budescu, and
  Wallsten]{turner2014forecast}
Brandon~M Turner, Mark Steyvers, Edgar~C Merkle, David~V Budescu, and Thomas~S
  Wallsten.
\newblock Forecast aggregation via recalibration.
\newblock \emph{Machine Learning}, 95\penalty0 (3):\penalty0 261--289, 2014.

\bibitem[Welinder et~al.(2013)Welinder, Welling, and Perona]{welinder2013lazy}
Peter Welinder, Max Welling, and Pietro Perona.
\newblock A lazy man's approach to benchmarking: Semisupervised classifier
  evaluation and recalibration.
\newblock In \emph{Proceedings of the IEEE Conference on Computer Vision and
  Pattern Recognition}, pages 3262--3269, 2013.

\bibitem[Wenger et~al.(2020)Wenger, Kjellstr{\"o}m, and Triebel]{wenger2020non}
Jonathan Wenger, Hedvig Kjellstr{\"o}m, and Rudolph Triebel.
\newblock Non-parametric calibration for classification.
\newblock In \emph{Proceedings of the International Conference on Artificial
  Intelligence and Statistics}, pages 178--190. PMLR, 2020.

\bibitem[Wick et~al.(2019)Wick, Panda, and Tristan]{wick2019unlocking}
Michael Wick, Swetasudha Panda, and Jean-Baptiste Tristan.
\newblock Unlocking fairness: a trade-off revisited.
\newblock In \emph{Advances in Neural Information Processing Systems}, pages
  8780--8789, 2019.

\bibitem[Zafar et~al.(2017)Zafar, Valera, Gomez~Rodriguez, and
  Gummadi]{zafar2017fairness}
Muhammad~Bilal Zafar, Isabel Valera, Manuel Gomez~Rodriguez, and Krishna~P
  Gummadi.
\newblock Fairness beyond disparate treatment \& disparate impact: Learning
  classification without disparate mistreatment.
\newblock In \emph{Proceedings of the 26th International Conference on World
  Wide Web}, pages 1171--1180, 2017.

\bibitem[Zhang et~al.(2020)Zhang, Li, Han, Zhou, Yu, et~al.]{zhang2020fairness}
Tao Zhang, Jing Li, Mengde Han, Wanlei Zhou, Philip Yu, et~al.
\newblock Fairness in semi-supervised learning: Unlabeled data help to reduce
  discrimination.
\newblock \emph{IEEE Transactions on Knowledge and Data Engineering}, 2020.

\end{thebibliography}

\clearpage
\beginsupplement

\title{Supplemental Material for ``Can I Trust My Fairness Metric? Assessing Fairness with Unlabeled Data and Bayesian Inference''}

\maketitle

\newpage
\addcontentsline{toc}{section}{Frequentist Estimation of Group Differences}
\section*{Appendix: Frequentist Estimation of Group Differences}

\begin{figure}[!h]
\centering 
   \subfloat{\includegraphics[width=0.5\linewidth]{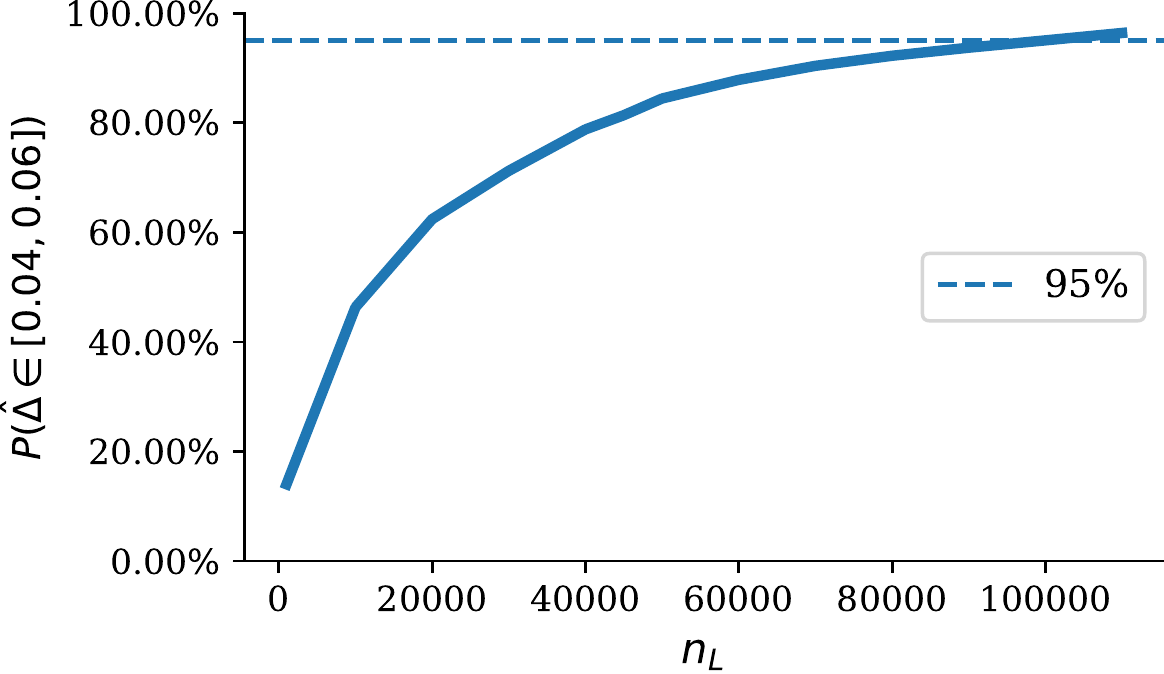}}
  \caption{Percentage of 10000 independent simulations whose estimates of $\Delta$ TPR are in the range $[0.04, 0.06]$, as a function of the  number of labeled examples $n_L$. } 
  \label{fig:amount_of_data}
\end{figure}

\paragraph{Synthetic Example (in Section 1):}
In Section 1 in the main paper we described a simple illustrative simulation to emphasize the point that large amounts of labeled data are often necessary to estimate groupwise fairness metrics accurately. The simulation consists of simulated data from two groups, where the underrepresented group makes up 20\% of the whole dataset, groupwise positive rates $P(y=1)$ are both 20\%, and the true groupwise TPRs are 95\% and  90\% (i.e., the true $\Delta$ is 0.05). TPR for group $g$ is defined as $P(\hat{y} = 1 | y = 1,  g)$ (See Section 2.1  in the paper for more details on notation). In Figure~\ref{fig:amount_of_data}, we show in this simulation that a large number $n_L$ of labeled examples (at least 96,000) is needed  to ensure there is a 95\% chance that our estimate of the true TPR difference (which is 0.05) lies in the range [0.04, 0.06].

\newpage
\addcontentsline{toc}{section}{Additional Details on Datasets and Classifiers}
\section*{Appendix: Additional Details on Datasets and Classifiers}

\paragraph{Datasets}
We performed experiments with six different real-world datasets. Summary statistics (e.g., test set size) are provided in Table 3 in the main paper. Below we provide additional details about these datasets in terms of background, relevant attributes, and how train and test splits were created. 
For all datasets, we preprocessed the data using the code from Friedler et al. (2018)\footnote{\url{https://github.com/algofairness/fairness-comparison/blob/master/fairness/preprocess.py}}.
We removed all instances that have missing data, and represented categorical variables with one-hot encoding. As in Friedler (2018), for all datasets except Adult we randomly sampled 2/3 of the data for training and use the remaining 1/3 for test. For the Adult data we re-split the training set of the original data into train and test as in Friedler et al. (2018).

\begin{itemize}
\item Adult: 
The Adult dataset\footnote{\url{https://archive.ics.uci.edu/ml/machine-learning-databases/adult}} from the  UCI Repository of Machine Learning Databases is based on   1994 U.S. census income data. 
This dataset consists of 14 demographic attributes for individuals. Instances are labeled according to whether their income exceeds \$50,000 per year.
In our experiments, ``race" and ``gender" are considered   sensitive attributes.
Instances are grouped into ``Amer-Indian-Inuit," ``Asian-Pac-Islander," ``Black," ``Other" and ``White" by race, and ``Female" and ``Male" by gender.
``White" and ``Male" are the privileged groups. 

\item Bank:
The Bank dataset\footnote{\url{http://archive.ics.uci.edu/ml/datasets/Bank+Marketing}} contains information about individual collected from a Portuguese banking institution. There are 20 attributes for each individuals, including marital status, education, and type of job.
The sensitive attribute we use is ``age," binarized by whether a individual's age is above 40 or not. The senior group is considered to be privileged.
Instances are labeled by whether the individual has subscribed to a term deposit account or not.

\item German: The German Credit dataset\footnote{\url{https://archive.ics.uci.edu/ml/machine-learning-databases/statlog/german}} from the UCI Repository of Machine Learning Databases describes individuals with 20 attributes including type of housing, credit history status, and employment status.
Each instance is labeled as being a good or bad credit risk.
The sensitive attributes used are ``gender" and ``age" (age at least 25 years old) and the privileged groups are defined as ``male" and ``adult."

\item Compas-R:
The ProPublica dataset\footnote{\url{https://github.com/propublica/compas-analysis}} contains information about the use of the COMPAS (Correctional Offender Management Profiling for Alternative Sanctions) risk assessment tool applied to 6,167 individuals in Broward County, Florida.
Each individual is labeled by whether they were rearrested within two years after the first arrest.
Sensitive attributees are ``gender" and ``race." 
By ``gender", individuals are grouped into ``Male" and ``Female"; by ``race", individuals are grouped into ``Caucasian." ``Asian," ``Native-American," ``African-American," ``Hispanic" and ``Others." The privileged groups are defined to be ``Male" and ``Caucasian." 

\item Compas-VR:
This is the violent recidivism version\footnote{\url{https://github.com/propublica/compas-analysis}} of the ProPublica data (Compas-R above), where the predicted outcome is a re-arrest for a violent crime.  

\item Ricci: 
The Ricci dataset\footnote{\url{https://ww2.amstat.org/publications/jse/v18n3/RicciData.csv}} is from the case of Ricci v. DeStefano from the Supreme Court of the United States (2009). It contains 118 instances and 5 attributes, including the sensitive attribute ``race." The privileged group was defined to be ``White."
Each instance is labeled by a promotion decision for each individual. 
\end{itemize}

\paragraph{Classification Models}
We used the following classification models in our experiments: logistic regression, multilayer perceptron (MLP) with a single hidden layer of size 10, random forests (the number of trees in the forest is set to 100), Gaussian Naive Bayes. The models were trained using standard default parameter settings and using the code provided by Friedler et al. (2018). Predictions from the trained models were  generated on the test data. Sensitive attributes were not included as inputs to the models during training or test.

\newpage
\addcontentsline{toc}{section}{Full Experimental Results}
\section*{Appendix: Complete Experimental Results}
In Figure 4 and in Tables 2 and 3 in the main paper we reported summary results of systematic comparisons between the frequentist method, the beta-binomial model (BB) method, and  the Bayesian calibration (BC) method, in terms of the mean absolute estimation error as a function of the number of labeled examples $n_L$.

In this section we provide complete tables and graphs for these results. In the tables the lowest error rate per row-column group is in bold if the difference among methods is statistically significant under a Wilcoxon signed-rank test (p=0.05).  As in the results in the main paper, the results below demonstrate that BC produces significantly more accurate estimates of group fairness metrics $\Delta$ than the BB or frequentist estimates, across all 4 classification models that we investigated, across FPR, TPR and Accuracy metrics, and across all datasets\footnote{``\textemdash'' in Tables~\ref{tab:tpr_full} and~\ref{tab:fpr_full} there are entries where the frequentist estimates of TPR or FPR do not exist.}

\begin{figure}[!h]
\centering
  \includegraphics[width=0.95\linewidth]{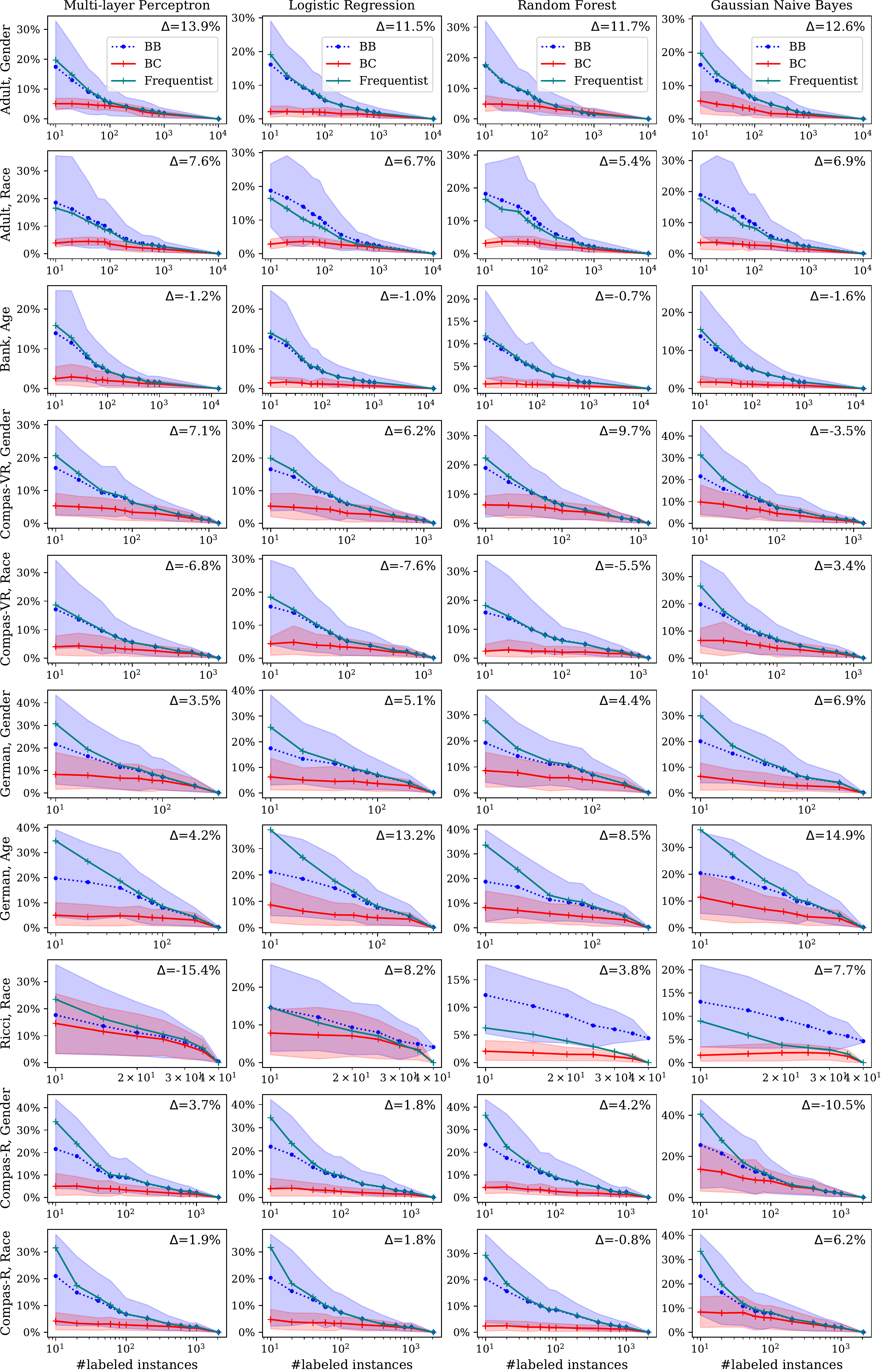}
  \caption{{\bf MAE for Accuracy:} Mean absolute error (MAE) of the difference between algorithm estimates and ground truth for group  difference in accuracy across 100 runs, as a function of number of labeled instances, for 10 different dataset-group pairs and 4 classifiers.  Shading indicates 95\% error bars for each method (not shown for the frequentist curve to avoid overplotting). Upper right corner shows the ground truth $\Delta$ between the unprivileged group and the privileged group.}
  \label{fig:MAE_accuracy}
\end{figure}

\begin{figure}[!h]
\centering
  \includegraphics[width=\linewidth]{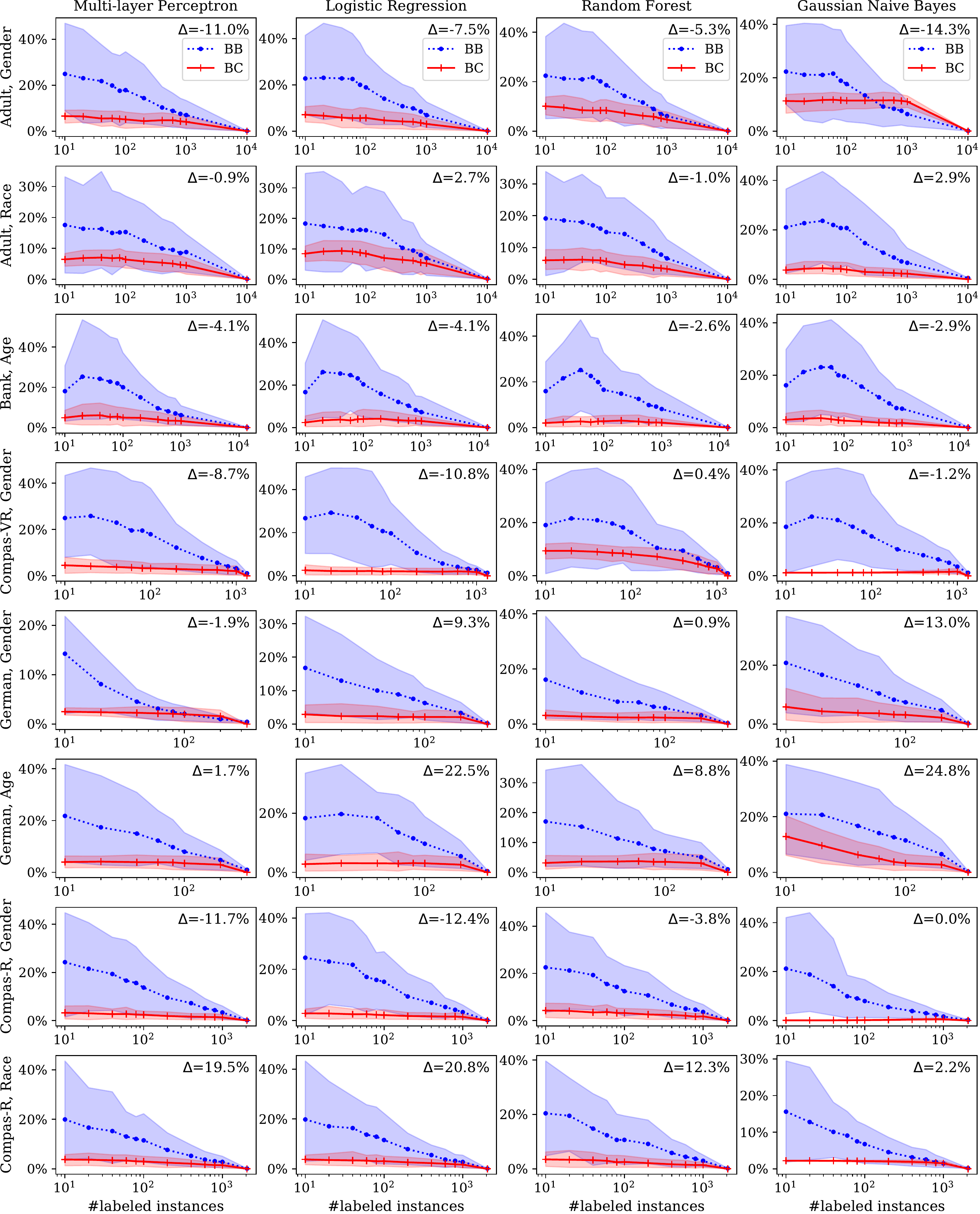}
  \caption{{\bf MAE for TPR:} Mean absolute error (MAE) of the difference between algorithm estimates and ground truth for group  difference in TPR across 100 runs. Compas-VR race and Ricci race are not included since there are no positive instances for some groups. Same setup as Figure~\ref{fig:MAE_accuracy}.}
  \label{fig:MAE_tpr}
\end{figure}

\begin{figure}[!h]
\centering
  \includegraphics[width=\linewidth]{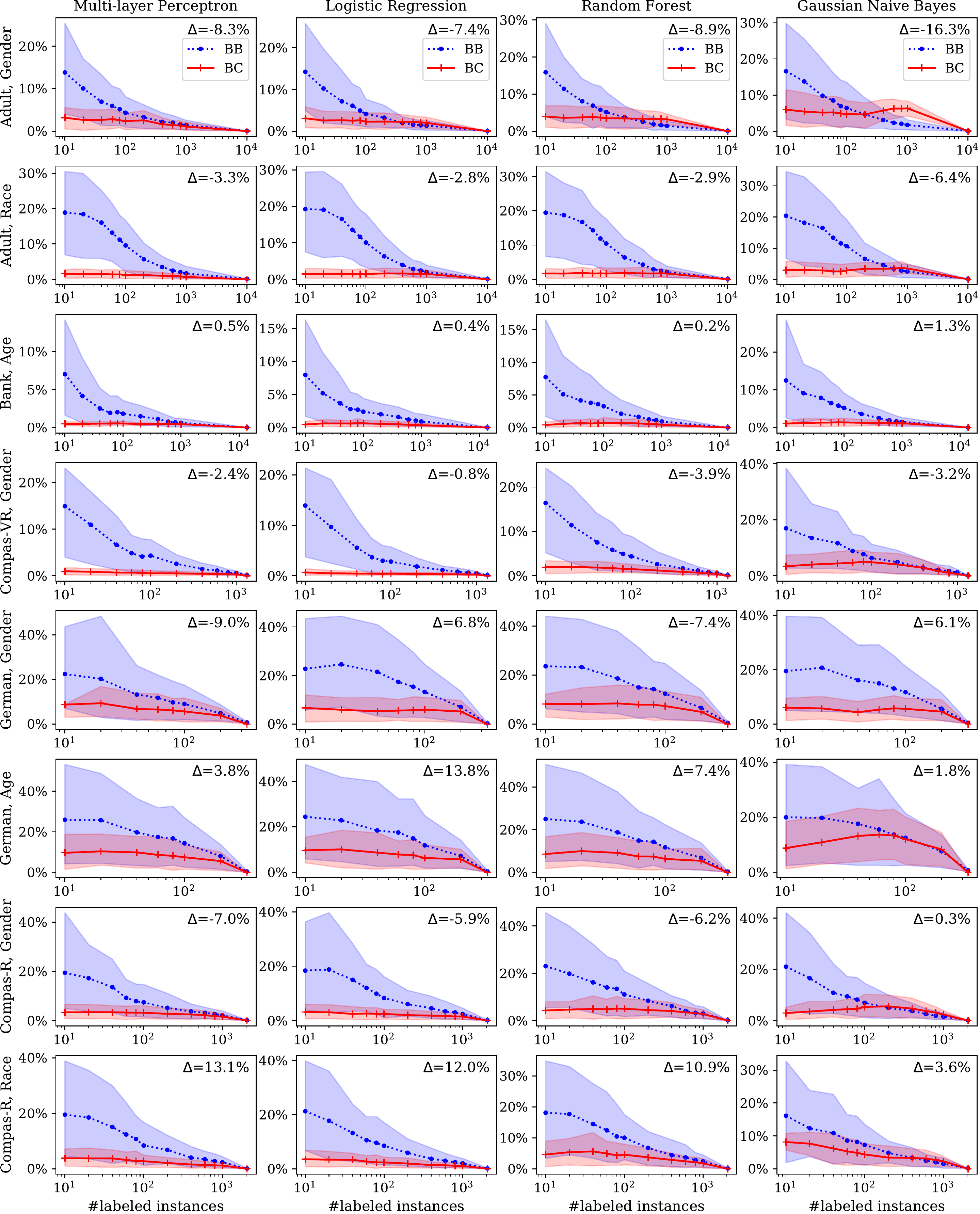}
  \caption{{\bf MAE for FPR:} Mean absolute error (MAE) of the difference between algorithm estimates and ground truth for groupwise difference in FPR across 100 runs. Same setup as Figure~\ref{fig:MAE_tpr}.}
  \label{fig:MAE_fpr}
\end{figure}

\begin{table}[!h]
\caption{{\bf MAE for $\Delta$  Accuracy Estimates}, with different $n_L$. Mean absolute error between estimates and true $\Delta$ across 100 runs of labeled samples of different sizes $n_L$ for different trained models (groups of columns) and 10 different dataset-group combinations (groups of rows). Estimation methods are Freq (frequentist), BB (beta-binomial), and BC (Bayesian-calibration). Freq and BB use only labeled samples, BC uses both labeled samples and unlabeled data.  Trained models are Multilayer Perceptron, Logistic Regression, Random Forests, and Gaussian NaiveBayes.
}
\resizebox{\textwidth}{!}{
\begin{tabular}{@{}rrrrrrcrrrcrrrcrrc@{}}
\toprule 
& 
& \phantom{a} &  \multicolumn{3}{c}{Multi-layer Perceptron}  
& \phantom{a}&  \multicolumn{3}{c}{Logistic Regression}
& \phantom{a} & \multicolumn{3}{c}{Random Forest} 
& \phantom{a} & \multicolumn{3}{c}{Gaussian Naive Bayes}\\ 
\cmidrule{4-6} \cmidrule{8-10} \cmidrule{12-14} \cmidrule{16-18}
Group & $n$ && Freq &BB  &BC && Freq &BB  &BC && Freq &BB  &BC && Freq &BB  &BC \\ \midrule
     Adult &  10 		&&16.5 &18.5 &\textbf{ 3.9}		&&16.4 &18.7 &\textbf{ 2.9}		&&16.5 &18.2 &\textbf{ 3.2}		&&17.6 &18.9 &\textbf{ 3.6}\\ 
      Race & 100 		&& 8.2 & 8.5 &\textbf{ 3.5}		&& 7.3 & 9.1 &\textbf{ 3.2}		&& 7.6 & 9.0 &\textbf{ 3.1}		&& 8.2 & 9.5 &\textbf{ 2.8}\\ 
           &1000 		&& 2.5 & 2.5 &\textbf{ 1.6}		&& 2.1 & 2.3 &\textbf{ 1.7}		&& 2.0 & 2.1 &\textbf{ 1.4}		&& 2.3 & 2.3 &\textbf{ 1.4}\\ 
\midrule 
     Adult &  10 		&&19.7 &17.4 &\textbf{ 5.1}		&&19.1 &16.1 &\textbf{ 2.2}		&&17.7 &17.4 &\textbf{ 4.8}		&&19.7 &16.2 &\textbf{ 5.4}\\ 
    Gender & 100 		&& 5.5 & 5.4 &4.4		&& 5.6 & 5.5 &\textbf{ 1.9}		&& 5.9 & 5.9 &\textbf{ 4.1}		&& 6.2 & 6.0 &\textbf{ 2.7}\\ 
           &1000 		&& 1.9 & 1.9 &1.6		&& 1.7 & 1.7 &\textbf{ 1.1}		&& 1.6 & \textbf{1.5} &2.0		&& 1.6 & 1.5 &\textbf{ 1.1}\\ 
\midrule 
      Bank &  10 		&&15.9 &13.9 &\textbf{ 2.5}		&&13.9 &13.0 &\textbf{ 1.4}		&&11.8 &11.1 &\textbf{ 1.0}		&&15.5 &13.7 &\textbf{ 1.7}\\ 
       Age & 100 		&& 4.4 & 4.3 &\textbf{ 2.0}		&& 4.3 & 4.3 &\textbf{ 1.2}		&& 4.3 & 4.2 &\textbf{ 0.9}		&& 5.0 & 5.0 &\textbf{ 1.1}\\ 
           &1000 		&& 1.5 & 1.5 &\textbf{ 1.1}		&& 1.6 & 1.6 &\textbf{ 0.7}		&& 1.4 & 1.4 &\textbf{ 0.5}		&& 1.7 & 1.7 &\textbf{ 0.8}\\ 
\midrule 
    German &  10 		&&34.6 &19.8 &\textbf{ 5.0}		&&37.1 &21.2 &\textbf{ 8.7}		&&33.6 &18.7 &\textbf{ 8.2}		&&36.6 &20.4 &\textbf{11.5}\\ 
       age & 100 		&& 8.5 & 8.0 &\textbf{ 3.9}		&& 8.2 & 7.6 &\textbf{ 3.8}		&& 8.8 & 8.2 &\textbf{ 4.3}		&& 9.7 & 9.1 &\textbf{ 4.2}\\ 
           & 200 		&& 4.4 & 4.2 &\textbf{ 3.1}		&& 4.5 & 4.4 &\textbf{ 3.3}		&& 4.9 & 4.8 &\textbf{ 3.3}		&& 4.8 & 4.7 &\textbf{ 3.5}\\ 
\midrule 
    German &  10 		&&30.7 &21.6 &\textbf{ 8.2}		&&25.6 &17.4 &\textbf{ 6.3}		&&27.7 &19.3 &\textbf{ 8.6}		&&30.0 &20.1 &\textbf{ 6.5}\\ 
    Gender & 100 		&& 7.3 & 7.1 &\textbf{ 5.4}		&& 7.1 & 6.9 &\textbf{ 3.7}		&& 7.2 & 7.0 &\textbf{ 4.8}		&& 6.0 & 5.9 &\textbf{ 2.8}\\ 
           & 200 		&& 3.2 & 3.2 &3.0		&& 4.0 & 3.9 &\textbf{ 2.9}		&& 3.6 & 3.5 &\textbf{ 2.9}		&& 4.0 & 4.0 &\textbf{ 2.2}\\ 
\midrule 
  Compas-R &  10 		&&31.5 &21.0 &\textbf{ 4.2}		&&31.7 &20.4 &\textbf{ 4.8}		&&29.3 &20.3 &\textbf{ 2.4}		&&33.5 &23.2 &\textbf{ 8.4}\\ 
      Race & 100 		&& 6.8 & 6.8 &\textbf{ 2.8}		&& 7.4 & 7.4 &\textbf{ 3.4}		&& 8.7 & 8.5 &\textbf{ 1.8}		&& 8.2 & 7.9 &\textbf{ 6.0}\\ 
           &1000 		&& 2.0 & 2.0 &\textbf{ 1.6}		&& 1.9 & 1.9 &1.6		&& 1.9 & 2.0 &\textbf{ 1.2}		&& 2.0 & 1.9 &1.8\\ 
\midrule 
  Compas-R &  10 		&&33.7 &21.6 &\textbf{ 5.0}		&&34.3 &21.9 &\textbf{ 3.8}		&&36.3 &23.3 &\textbf{ 4.4}		&&40.5 &25.5 &\textbf{13.7}\\ 
    Gender & 100 		&& 9.3 & 8.8 &\textbf{ 3.3}		&& 9.5 & 9.0 &\textbf{ 2.6}		&& 8.8 & 8.5 &\textbf{ 2.7}		&&10.2 & 9.7 &\textbf{ 8.0}\\ 
           &1000 		&& 2.1 & 2.0 &\textbf{ 1.4}		&& 2.2 & 2.2 &\textbf{ 1.3}		&& 2.4 & 2.4 &\textbf{ 1.4}		&& 1.9 & 1.9 &\textbf{ 1.8}\\ 
\midrule 
 Compas-VR &  10 		&&18.7 &17.1 &\textbf{ 4.0}		&&18.5 &15.6 &\textbf{ 4.4}		&&18.2 &15.8 &\textbf{ 2.4}		&&26.6 &19.8 &\textbf{ 6.5}\\ 
      Race & 100 		&& 5.5 & 5.6 &\textbf{ 3.1}		&& 5.1 & 5.1 &\textbf{ 3.4}		&& 6.0 & 6.3 &\textbf{ 2.0}		&& 6.8 & 6.6 &\textbf{ 3.7}\\ 
           &1000 		&& 0.9 & 0.9 &\textbf{ 0.8}		&& 0.9 & 0.9 &0.8		&& 0.9 & 0.9 &\textbf{ 0.8}		&& 1.1 & 1.1 &\textbf{ 0.9}\\ 
\midrule 
 Compas-VR &  10 		&&20.6 &16.9 &\textbf{ 5.4}		&&19.9 &16.6 &\textbf{ 5.3}		&&22.3 &19.0 &\textbf{ 6.3}		&&31.3 &21.5 &\textbf{ 9.8}\\ 
    Gender & 100 		&& 6.4 & 6.3 &\textbf{ 3.4}		&& 6.1 & 6.0 &\textbf{ 3.1}		&& 6.3 & 6.3 &\textbf{ 4.4}		&& 7.3 & 7.1 &\textbf{ 4.5}\\ 
           &1000 		&& 1.0 & 1.0 &\textbf{ 0.9}		&& 1.0 & 1.0 &\textbf{ 0.9}		&& 0.9 & 0.9 &1.0		&& 1.4 & 1.4 &\textbf{ 0.9}\\ 
\midrule 
     Ricci &  10 		&&23.5 &17.7 &\textbf{14.6}		&&14.6 &14.6 &\textbf{ 7.9}		&& 6.3 &12.2 &\textbf{ 2.1}		&& 8.9 &13.1 &\textbf{ 1.6}\\ 
      Race &  20 		&&12.9 &11.1 &9.8		&& 8.4 & 9.3 &\textbf{ 7.1}		&& 3.9 & 8.5 &\textbf{ 1.5}		&& 3.8 & 9.4 &\textbf{ 2.1}\\ 
           &  30 		&& 8.5 & 7.5 &\textbf{ 6.5}		&& 4.9 & 5.7 &\textbf{ 4.6}		&& 2.0 & 6.0 &\textbf{ 1.1}		&& 2.8 & 6.5 &\textbf{ 2.0}\\ 
\bottomrule
\end{tabular}
\label{tab:acc_full}
}
\end{table}

\begin{table}[h]
\caption{{\bf MAE for $\Delta$  TPR Estimates}, with different $n_L$. Mean absolute error between estimates and true $\Delta$ across 100 runs of labeled samples of different sizes $n_L$ for different trained models (groups of columns) and 8 different dataset-group combinations (groups of rows).   Estimation methods are Freq (Frequentist), BB (beta-binomial), and BC (Bayesian-calibration). Freq and BB use only labeled samples, BC uses both labeled samples and unlabeled data. Trained models are Multilayer Perceptron, Logistic Regression, Random Forests, and Gaussian NaiveBayes.
}
\resizebox{\textwidth}{!}{%
\begin{tabular}{@{}rrrrrrcrrrcrrrcrrc@{}}
\toprule 
& 
& \phantom{a} &  \multicolumn{3}{c}{Multi-layer Perceptron}  
& \phantom{a}&  \multicolumn{3}{c}{Logistic Regression}
& \phantom{a} & \multicolumn{3}{c}{Random Forest} 
& \phantom{a} & \multicolumn{3}{c}{Gaussian Naive Bayes}\\ 
\cmidrule{4-6} \cmidrule{8-10} \cmidrule{12-14} \cmidrule{16-18}
Group & $n$ && Freq &BB  &BC && Freq &BB  &BC && Freq &BB  &BC && Freq &BB  &BC \\ \midrule
    Adult &  40 		&&\textemdash&16.3 &\textbf{ 7.0}		&&\textemdash&16.7 &\textbf{ 9.3}		&&\textemdash&17.9 &\textbf{ 6.2}		&&\textemdash&23.6 &\textbf{ 4.5}\\ 
     Race & 100 		&&\textemdash&15.3 &\textbf{ 6.4}		&&\textemdash&16.1 &\textbf{ 8.4}		&&\textemdash&14.9 &\textbf{ 5.6}		&&\textemdash&20.7 &\textbf{ 3.9}\\ 
          & 200 		&&\textemdash&12.5 &\textbf{ 5.8}		&&\textemdash&14.7 &\textbf{ 7.0}		&&\textemdash&14.3 &\textbf{ 4.6}		&&\textemdash&14.6 &\textbf{ 3.0}\\ 
\midrule 
    Adult &  40 		&&\textemdash&21.8 &\textbf{ 5.5}		&&\textemdash&22.8 &\textbf{ 5.8}		&&\textemdash&20.9 &\textbf{ 8.4}		&&\textemdash&21.1 &\textbf{11.7}\\ 
   Gender & 100 		&&\textemdash&17.8 &\textbf{ 5.1}		&&\textemdash&18.9 &\textbf{ 5.7}		&&\textemdash&18.6 &\textbf{ 8.4}		&&\textemdash&17.7 &\textbf{11.4}\\ 
          & 200 		&&16.3 &14.3 &\textbf{ 4.3}		&&15.8 &14.0 &\textbf{ 4.6}		&&16.1 &14.2 &\textbf{ 7.3}		&&15.0 &13.4 &11.5\\ 
\midrule 
     Bank &  40 		&&\textemdash&24.2 &\textbf{ 6.1}		&&\textemdash&25.4 &\textbf{ 3.8}		&&\textemdash&25.2 &\textbf{ 2.7}		&&\textemdash&23.0 &\textbf{ 3.6}\\ 
      Age & 100 		&&25.9 &20.0 &\textbf{ 5.0}		&&25.7 &20.4 &\textbf{ 4.0}		&&20.9 &16.6 &\textbf{ 2.8}		&&24.9 &19.6 &\textbf{ 2.6}\\ 
          & 200 		&&16.8 &15.0 &\textbf{ 4.8}		&&17.7 &15.9 &\textbf{ 4.2}		&&16.6 &14.9 &\textbf{ 3.1}		&&17.3 &15.7 &\textbf{ 2.3}\\ 
\midrule 
   German &  40 		&&\textemdash&15.0 &\textbf{ 3.9}		&&\textemdash&18.4 &\textbf{ 3.0}		&&\textemdash&11.3 &\textbf{ 3.6}		&&\textemdash&16.7 &\textbf{ 6.3}\\ 
      age & 100 		&& 8.9 & 8.0 &\textbf{ 3.5}		&&10.7 & 9.7 &\textbf{ 3.1}		&& 8.0 & 7.1 &\textbf{ 3.5}		&&12.9 &11.5 &\textbf{ 3.3}\\ 
          & 200 		&& 4.7 & 4.7 &\textbf{ 3.0}		&& 5.6 & 5.4 &\textbf{ 2.6}		&& 5.1 & 5.1 &\textbf{ 3.1}		&& 6.8 & 6.5 &\textbf{ 2.8}\\ 
\midrule 
   German &  40 		&& 2.6 & 4.5 &\textbf{ 2.3}		&&11.8 &10.0 &\textbf{ 2.4}		&& 9.4 & 8.1 &\textbf{ 2.4}		&&15.0 &13.1 &\textbf{ 3.8}\\ 
   Gender & 100 		&& 1.4 & 2.1 &2.0		&& 6.5 & 6.3 &\textbf{ 2.1}		&& 5.9 & 5.8 &\textbf{ 2.3}		&& 7.7 & 7.4 &\textbf{ 3.1}\\ 
          & 200 		&& \textbf{0.7 }& 1.0 & 1.6		&& 3.3 & 3.3 &\textbf{ 2.1}		&& 3.1 & 3.2 &\textbf{ 2.1}		&& 4.8 & 4.7 &\textbf{ 2.2}\\ 
\midrule 
 Compas-R &  40 		&&\textemdash&15.2 &\textbf{ 3.4}		&&\textemdash&16.3 &\textbf{ 3.4}		&&\textemdash&14.8 &\textbf{ 3.2}		&&\textemdash&10.1 &\textbf{ 2.2}\\ 
     Race & 100 		&&\textemdash&11.5 &\textbf{ 2.9}		&&\textemdash&11.5 &\textbf{ 3.1}		&&\textemdash&10.6 &\textbf{ 2.5}		&&\textemdash& 6.7 &\textbf{ 2.1}\\ 
          & 200 		&&\textemdash& 7.6 &\textbf{ 2.5}		&&\textemdash& 7.9 &\textbf{ 2.6}		&&\textemdash& 9.2 &\textbf{ 2.1}		&&\textemdash& 4.5 &\textbf{ 2.0}\\ 
\midrule 
 Compas-R &  40 		&&\textemdash&19.3 &\textbf{ 2.7}		&&\textemdash&21.8 &\textbf{ 2.5}		&&\textemdash&19.3 &\textbf{ 3.4}		&&\textemdash&14.0 &\textbf{ 0.1}\\ 
   Gender & 100 		&&15.9 &13.7 &\textbf{ 2.4}		&&17.6 &15.1 &\textbf{ 2.1}		&&14.3 &12.5 &\textbf{ 3.2}		&& 8.7 & 8.0 &\textbf{ 0.2}\\ 
          & 200 		&&10.0 & 9.5 &\textbf{ 1.9}		&&10.0 & 9.4 &\textbf{ 1.8}		&&11.3 &10.7 &\textbf{ 2.6}		&& 5.6 & 5.5 &\textbf{ 0.3}\\ 
\midrule 
Compas-VR &  40 		&&\textemdash&23.0 &\textbf{ 3.8}		&&\textemdash&27.0 &\textbf{ 2.2}		&&\textemdash&20.9 &\textbf{ 9.0}		&&\textemdash&21.1 &\textbf{ 1.2}\\ 
   Gender & 100 		&&\textemdash&18.0 &\textbf{ 3.2}		&&\textemdash&19.7 &\textbf{ 2.1}		&&\textemdash&16.3 &\textbf{ 8.1}		&&\textemdash&14.9 &\textbf{ 1.2}\\ 
          & 200 		&&14.9 &12.2 &\textbf{ 2.9}		&& 8.9 &10.7 &\textbf{ 2.0}		&&14.6 &10.5 &7.2		&&12.5 &10.0 &\textbf{ 1.3}\\ 
\bottomrule
\end{tabular}
\label{tab:tpr_full}
}
\end{table}

\begin{table}[h]
\caption{{\bf MAE for $\Delta$ FPR Estimates}, with different $n_L$. Same setup as Table~\ref{tab:tpr_full}.
}
\resizebox{\textwidth}{!}{
\begin{tabular}{@{}rrrrrrcrrrcrrrcrrc@{}}
\toprule 
& 
& \phantom{a} &  \multicolumn{3}{c}{Multi-layer Perceptron}  
& \phantom{a}&  \multicolumn{3}{c}{Logistic Regression}
& \phantom{a} & \multicolumn{3}{c}{Random Forest} 
& \phantom{a} & \multicolumn{3}{c}{Gaussian Naive Bayes}\\ 
\cmidrule{4-6} \cmidrule{8-10} \cmidrule{12-14} \cmidrule{16-18}
Group & $n$ && Freq &BB  &BC && Freq &BB  &BC && Freq &BB  &BC && Freq &BB  &BC \\ \midrule
    Adult &  40 		&&\textemdash&16.1 &\textbf{ 1.5}		&&\textemdash&16.6 &\textbf{ 1.5}		&&\textemdash&16.7 &\textbf{ 1.8}		&&\textemdash&16.5 &\textbf{ 2.9}\\ 
     Race & 100 		&&\textemdash& 9.6 &\textbf{ 1.2}		&&\textemdash&10.1 &\textbf{ 1.5}		&&\textemdash&10.5 &\textbf{ 1.7}		&&\textemdash&10.7 &\textbf{ 2.8}\\ 
          & 200 		&&\textemdash& 5.7 &\textbf{ 1.2}		&&\textemdash& 6.3 &\textbf{ 1.6}		&&\textemdash& 6.4 &\textbf{ 1.8}		&&\textemdash& 6.6 &\textbf{ 3.4}\\ 
\midrule 
    Adult &  40 		&& 7.1 & 6.9 &\textbf{ 2.6}		&& 7.2 & 7.1 &\textbf{ 2.6}		&& 8.3 & 8.1 &\textbf{ 3.7}		&&10.3 & 9.8 &\textbf{ 5.1}\\ 
   Gender & 100 		&& 4.4 & 4.3 &\textbf{ 2.3}		&& 4.3 & 4.1 &\textbf{ 2.2}		&& 5.2 & 5.1 &\textbf{ 3.5}		&& 6.6 & 6.4 &\textbf{ 4.7}\\ 
          & 200 		&& 3.2 & 3.3 &\textbf{ 2.5}		&& 3.2 & 3.2 &\textbf{ 2.3}		&& 3.7 & 3.7 &\textbf{ 3.4}		&& 4.7 & 4.6 &\textbf{ 4.6}\\ 
\midrule 
     Bank &  40 		&& 2.4 & 2.5 &\textbf{ 0.5}		&& 3.6 & 3.7 &\textbf{ 0.6}		&& 4.1 & 4.2 &\textbf{ 0.7}		&& 8.5 & 7.9 &\textbf{ 1.3}\\ 
      Age & 100 		&& 1.9 & 1.8 &\textbf{ 0.5}		&& 2.4 & 2.4 &\textbf{ 0.6}		&& 3.3 & 3.3 &\textbf{ 0.7}		&& 5.3 & 5.2 &\textbf{ 1.3}\\ 
          & 200 		&& 1.5 & 1.5 &\textbf{ 0.5}		&& 2.1 & 2.0 &\textbf{ 0.6}		&& 2.1 & 2.1 &\textbf{ 0.7}		&& 3.6 & 3.6 &\textbf{ 1.3}\\ 
\midrule 
   German &  40 		&&\textemdash&19.7 &\textbf{ 9.8}		&&\textemdash&18.4 &\textbf{ 8.7}		&&\textemdash&18.7 &\textbf{ 9.1}		&&\textemdash&17.6 &\textbf{13.2}\\ 
      age & 100 		&&16.6 &14.3 &\textbf{ 7.4}		&&13.6 &11.9 &\textbf{ 6.3}		&&13.7 &11.7 &\textbf{ 6.3}		&&14.9 &12.5 &\textbf{12.0}\\ 
          & 200 		&& 8.6 & 8.0 &\textbf{ 5.6}		&& 7.7 & 7.2 &\textbf{ 5.7}		&& 7.2 & 6.8 &\textbf{ 5.4}		&& 8.4 & \textbf{7.7} & 8.3\\ 
\midrule 
   German &  40 		&&15.6 &13.2 &\textbf{ 6.8}		&&27.3 &21.5 &\textbf{ 5.3}		&&23.1 &18.6 &\textbf{ 8.4}		&&20.3 &16.1 &\textbf{ 4.4}\\ 
   Gender & 100 		&& 9.2 & 9.0 &\textbf{ 5.7}		&&14.4 &13.2 &\textbf{ 5.9}		&&13.3 &12.4 &\textbf{ 7.4}		&&12.6 &11.7 &\textbf{ 5.6}\\ 
          & 200 		&& 4.9 & 4.9 &\textbf{ 3.8}		&& 7.3 & 7.0 &\textbf{ 5.2}		&& 6.8 & 6.6 &\textbf{ 5.0}		&& 5.9 & 5.7 &\textbf{ 4.6}\\ 
\midrule 
 Compas-R &  40 		&&\textemdash&15.1 &\textbf{ 3.7}		&&\textemdash&13.2 &\textbf{ 3.3}		&&\textemdash&14.5 &\textbf{ 5.6}		&&\textemdash&10.8 &\textbf{ 6.2}\\ 
     Race & 100 		&&\textemdash& 8.4 &\textbf{ 2.7}		&&\textemdash& 8.5 &\textbf{ 2.4}		&&\textemdash&10.0 &\textbf{ 4.6}		&&\textemdash& 7.3 &\textbf{ 4.4}\\ 
          & 200 		&&\textemdash& 6.8 &\textbf{ 2.1}		&&\textemdash& 5.9 &\textbf{ 1.9}		&&\textemdash& 6.7 &\textbf{ 3.7}		&&\textemdash& 4.9 &\textbf{ 3.4}\\ 
\midrule 
 Compas-R &  40 		&&\textemdash&13.5 &\textbf{ 3.4}		&&\textemdash&15.0 &\textbf{ 2.4}		&&\textemdash&16.2 &\textbf{ 4.9}		&&\textemdash&10.9 &\textbf{ 4.2}\\ 
   Gender & 100 		&& 7.7 & 7.4 &\textbf{ 3.2}		&& 8.5 & 8.3 &\textbf{ 2.4}		&&11.5 &11.0 &\textbf{ 5.0}		&& 7.4 & 6.9 &\textbf{ 5.3}\\ 
          & 200 		&& 5.3 & 5.2 &\textbf{ 2.7}		&& 6.1 & 6.1 &\textbf{ 2.0}		&& 8.5 & 8.4 &\textbf{ 4.4}		&& 5.1 & \textbf{5.0} & 5.6\\ 
\midrule 
Compas-VR &  40 		&& 5.6 & 6.6 &\textbf{ 0.7}		&& 3.3 & 5.6 &\textbf{ 0.4}		&& 5.5 & 7.5 &\textbf{ 1.9}		&&12.8 &11.7 &\textbf{ 4.4}\\ 
   Gender & 100 		&& 4.0 & 4.3 &\textbf{ 0.6}		&& 2.4 & 2.8 &\textbf{ 0.4}		&& 3.9 & 4.4 &\textbf{ 1.5}		&& 6.3 & 6.3 &\textbf{ 4.8}\\ 
          & 200 		&& 2.5 & 2.6 &\textbf{ 0.5}		&& 1.8 & 1.8 &\textbf{ 0.4}		&& 2.5 & 2.6 &\textbf{ 1.2}		&& 5.1 & 4.9 &\textbf{ 4.1}\\ 
\bottomrule
\end{tabular}
\label{tab:fpr_full}
}
\end{table}

\clearpage
\newpage
\addcontentsline{toc}{section}{Calibration Coverage of Posterior Credible Intervals}
\section*{Appendix: Calibration Coverage of Posterior Credible Intervals}

We can generate posterior credible intervals on $\Delta$ (as shown in red in Figure 3 in the main paper) for both the BB and BC methods by computing upper and lower percentiles from posterior samples for $\Delta$. Below in Table~\ref{tab:coverage} we show
the coverage of   95\% credible intervals for both the BB (beta-bernoulli) and BC (Bayesian-calibration) methods, for the multi-layer perceptron model. Coverage is defined as the percentage of credible intervals (across multiple different labeled datasets of size $n_L$) that contain the true value: a perfectly calibrated 95\% credible interval would have 95\% coverage.  Table~\ref{tab:coverage} shows that while the coverage for both methods is generally not far from 95\% there is room for improvement (as discussed in the main paper). For example, for small values of $n_L$ the coverage of both methods is often too high (above 95\%), with some evidence of coverage decreasing as $n_L$ increasing. Generating accurate posterior credible intervals is a known issue in Bayesian analysis in  the presence of model misspecification (e.g., Syring and Martin (2019)) and is an interesting direction for future work on Bayesian analysis of fairness metrics.

\begin{table}[!h]
\caption{\textbf{Calibration Coverage of Posterior Credible Intervals Comparison}, across 1000 runs of labeled samples of different sizes $n_L$ for 10 different dataset-group combinations (rows). Estimation methods are BC (Bayesian-Calibration) and BB (beta-bernoulli). Trained model is Multilayer Perceptron.}
\centering
\resizebox{0.85\textwidth}{!}{%
\begin{tabular}{@{}rrrccccccccccc@{}}
\toprule 
& 
& \phantom{a} &  \multicolumn{2}{c}{$n_L=10$}
& \phantom{a} &  \multicolumn{2}{c}{$n_L=20$}
& \phantom{a} &  \multicolumn{2}{c}{$n_L=40$}
& \phantom{a} &  \multicolumn{2}{c}{$n_L=100$}\\ 
\cmidrule{4-5} \cmidrule{7-8} \cmidrule{10-11} \cmidrule{13-14}
\multicolumn{2}{c}{Group}  && BC  &BB && BC  &BB && BC  &BB && BC  &BB \\ \midrule
\multicolumn{2}{c}{Adult, Race}	&&99.9 &97.7 &&98.6 &93.5 &&96.2 &93.2 &&92.3 &95.3 \\ 
\multicolumn{2}{c}{Adult, Gender}	&&100.0 &96.4 &&99.7 &95.5 &&99.2 &94.9 &&96.8 &95.5 \\ 
\multicolumn{2}{c}{ Bank, Age}	&&99.4 &98.7 &&98.8 &98.5 &&98.0 &96.4 &&93.7 &95.3 \\ 
\multicolumn{2}{c}{German, age}	&&99.9 &98.8 &&99.6 &98.1 &&99.0 &98.3 &&96.9 &98.3 \\ 
\multicolumn{2}{c}{German, Gender}	&&99.1 &97.4 &&99.1 &97.4 &&97.7 &96.4 &&94.6 &97.8 \\ 
\multicolumn{2}{c}{Compas-R, Race}	&&99.3 &98.8 &&99.4 &97.2 &&99.1 &96.7 &&99.3 &96.6 \\ 
\multicolumn{2}{c}{Compas-R, Gender}	&&99.3 &97.7 &&99.3 &97.0 &&98.6 &95.9 &&97.6 &96.5 \\ 
\multicolumn{2}{c}{Compas-VR, Race}	&&99.6 &100.0 &&98.6 &97.8 &&97.9 &95.2 &&97.5 &93.1 \\ 
\multicolumn{2}{c}{Compas-VR, Gender}	&&96.3 &97.2 &&94.3 &96.5 &&95.4 &96.1 &&95.8 &97.1 \\ 
\multicolumn{2}{c}{Ricci, Race}	&&93.2 &99.7 &&91.4 &99.7 && \textemdash &\textemdash && \textemdash & \textemdash \\ 
\bottomrule
\end{tabular}
}
\label{tab:coverage}
\end{table}

\clearpage
\newpage
\addcontentsline{toc}{section}{Graphical Model for Hierarchical Beta Calibration}
\section*{Appendix: Graphical Model for Hierarchical Beta Calibration}
\begin{figure}[!h]
\centering
  \includegraphics[width=0.6\linewidth]{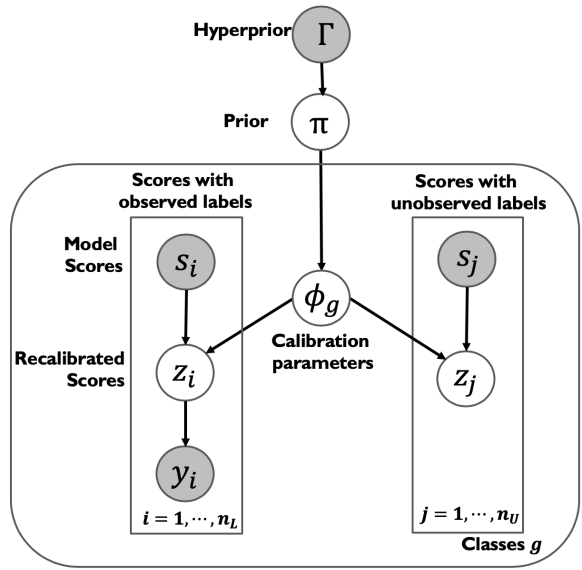}
  \caption{Graphical model for hierarchical beta calibration as described in Section 2.4 of the main paper. $\Gamma$ is the hyperprior on $\pi$, representing the fixed parameters for the normal and truncated normal hyperpriors described in Section 2.4 in the main paper.}
  \label{fig:diagram}
\end{figure}

\clearpage
\newpage
\addcontentsline{toc}{section}{Error Results with with Non-Hierarchical Bayesian Calibration}
\section*{Appendix: Error Results with with Non-Hierarchical Bayesian Calibration}
In our Hierarchical Bayesian calibration model we allows different groups to share statistical strength via a hierarchical structure. In this section, we compare our proposed Bayesian calibration model (BC) that uses this hierarchy with a non-hierarchical Bayesian calibration (NHBC) approach.
Table~\ref{tab:ablation} compares the mean absolute error (MAE) rate for both approaches in estimating differences in accuracy between groups (same setup as Tables 2 and 3 in the main paper in terms of how MAE is computed). The results show that (1) both BC and NHBC significantly improve MAE compared to BB; (2) BC and NHBC are comparable in most  cases, but with the hierarchical structure the BC method avoids occasional catastrophic errors that NHBC can make, e.g. when assessing $\Delta$ Accuracy of a Gaussian Naive Bayes model on Compas-R Gender and Compas-VR Gender.

\begin{table}[!h]
\caption{{\bf MAE for $\Delta$  Accuracy Estimates}, with different $n_L$. Mean absolute error between estimates and true $\Delta$ across 100 runs of labeled samples of different sizes $n_L$ for different trained models (groups of columns) and 10 different dataset-group combinations (groups of rows). Estimation methods are BB (beta-binomial), and NHBC (non-hierarchical Bayesian-calibration), BC (Bayesian-Calibration). BB uses only labeled samples, NHBC and BC use both labeled samples and unlabeled data.  Trained models are Multilayer Perceptron, Logistic Regression, Random Forests, and Gaussian NaiveBayes.}
\centering
\resizebox{\textwidth}{!}{%
\begin{tabular}{@{}rrrrrrcrrrcrrrcrrr@{}}
\toprule 
& 
& \phantom{a} &  \multicolumn{3}{c}{Multi-layer Perceptron}  
& \phantom{a}&  \multicolumn{3}{c}{Logistic Regression}
& \phantom{a} & \multicolumn{3}{c}{Random Forest} 
& \phantom{a} & \multicolumn{3}{c}{Gaussian Naive Bayes}\\ 
\cmidrule{4-6} \cmidrule{8-10} \cmidrule{12-14} \cmidrule{16-18}
Group & $n$ && BB  &NHBC &BC && BB  &NHBC &BC && BB  &NHBC &BC && BB  &NHBC &BC \\ \midrule
     Adult &  10 		&&18.4  &\textbf{ 3.2} & 3.9  			&&18.8  &\textbf{ 2.7} & 2.9  			&&18.1  &\textbf{ 2.8} & 3.2  			&&18.9  & 4.5  &\textbf{ 3.6} 	\\ 
      Race &  20 		&&16.1  &\textbf{ 3.3} & 4.4  			&&16.7  &\textbf{ 2.9} & 3.4  			&&16.3  &\textbf{ 3.0} & 3.7  			&&16.8  & 4.1  &\textbf{ 3.7} 	\\ 
           &  40 		&&13.1  &\textbf{ 2.8} & 4.5  			&&14.0  &\textbf{ 2.9} & 3.7  			&&14.4  &\textbf{ 2.9} & 3.8  			&&14.4  & 3.7  &\textbf{ 3.3} 	\\ 
           & 100 		&& 8.6  &\textbf{ 2.7} & 3.5  			&& 9.2  &\textbf{ 3.0} & 3.2  			&& 9.0  &\textbf{ 2.6} & 3.1  			&& 9.6  &\textbf{ 2.4} & 2.8  	\\ 
           &1000 		&& 2.5  &\textbf{ 1.4} & 1.6  			&& 2.3  & 2.1  &\textbf{ 1.7} 			&& 2.1  &\textbf{ 0.7} & 1.4  			&& 2.3  & 1.8  &\textbf{ 1.4} 	\\ 
\midrule 
     Adult &  10 		&&17.4  &\textbf{ 4.1} & 5.1  			&&16.3  & 2.6  &\textbf{ 2.2} 			&&17.3  & 5.3  &\textbf{ 4.8} 			&&16.3  & 7.2  &\textbf{ 5.4} 	\\ 
    Gender &  20 		&&12.9  &\textbf{ 4.4} & 5.1  			&&12.2  & 2.6  &\textbf{ 2.2} 			&&12.4  & 5.3  &\textbf{ 4.9} 			&&11.6  & 6.7  &\textbf{ 4.5} 	\\ 
           &  40 		&& 9.0  &\textbf{ 4.1} & 4.9  			&& 9.2  & 2.5  &\textbf{ 2.1} 			&& 9.6  & 5.1  &\textbf{ 4.5} 			&& 9.7  & 6.3  &\textbf{ 3.9} 	\\ 
           & 100 		&& 5.4  &\textbf{ 3.1} & 4.4  			&& 5.5  & 2.0  &\textbf{ 2.0} 			&& 5.9  & 4.7  &\textbf{ 4.1} 			&& 6.0  & 4.8  &\textbf{ 2.7} 	\\ 
           &1000 		&& 1.9  &\textbf{ 1.4} & 1.6  			&& 1.7  &\textbf{ 1.0} & 1.1  			&&\textbf{ 1.5} & 1.8  & 2.0  			&& 1.5  &\textbf{ 0.9} & 1.0  	\\ 
\midrule 
      Bank &  10 		&&14.0  &\textbf{ 1.7} & 2.5  			&&12.8  & 1.5  &\textbf{ 1.4} 			&&11.2  & 1.1  &\textbf{ 1.0} 			&&13.7  &\textbf{ 1.4} & 1.7  	\\ 
       Age &  20 		&&11.6  &\textbf{ 2.3} & 2.9  			&&10.9  & 1.9  &\textbf{ 1.7} 			&& 8.8  & 1.4  &\textbf{ 1.2} 			&&10.3  &\textbf{ 1.6} & 1.7  	\\ 
           &  40 		&& 8.0  &\textbf{ 2.3} & 2.6  			&& 7.3  & 1.7  &\textbf{ 1.4} 			&& 6.5  & 1.5  &\textbf{ 1.1} 			&& 7.5  & 1.7  &\textbf{ 1.5} 	\\ 
           & 100 		&& 4.3  & 2.2  &\textbf{ 2.0} 			&& 4.3  & 1.4  &\textbf{ 1.2} 			&& 4.2  & 1.2  &\textbf{ 0.9} 			&& 4.9  & 1.3  &\textbf{ 1.1} 	\\ 
           &1000 		&& 1.5  & 1.2  &\textbf{ 1.1} 			&& 1.6  & 0.8  &\textbf{ 0.7} 			&& 1.4  & 0.6  &\textbf{ 0.5} 			&& 1.7  &\textbf{ 0.7} & 0.8  	\\ 
\midrule 
    German &  10 		&&19.7  & 5.6  &\textbf{ 5.0} 			&&21.3  &10.3  &\textbf{ 8.7} 			&&19.1  &\textbf{ 8.2} & 8.2  			&&20.4  &14.2  &\textbf{11.5} 	\\ 
       age &  20 		&&18.1  & 6.0  &\textbf{ 4.4} 			&&18.6  & 6.7  &\textbf{ 6.4} 			&&16.7  &\textbf{ 7.0} & 7.0  			&&18.8  & 9.9  &\textbf{ 9.0} 	\\ 
           &  40 		&&15.9  & 6.7  &\textbf{ 4.8} 			&&15.0  & 5.6  &\textbf{ 4.9} 			&&11.7  & 6.6  &\textbf{ 5.8} 			&&14.9  &\textbf{ 6.4} & 6.9  	\\ 
           & 100 		&& 7.9  & 5.8  &\textbf{ 3.9} 			&& 7.5  & 5.5  &\textbf{ 3.8} 			&& 8.2  & 6.5  &\textbf{ 4.3} 			&& 9.1  & 4.4  &\textbf{ 4.2} 	\\ 
           & 200 		&& 4.2  & 3.7  &\textbf{ 3.1} 			&& 4.4  & 4.1  &\textbf{ 3.3} 			&& 4.7  & 4.1  &\textbf{ 3.3} 			&& 4.7  & 3.8  &\textbf{ 3.5} 	\\ 
\midrule 
    German &  10 		&&21.5  &10.5  &\textbf{ 8.2} 			&&17.6  & 7.0  &\textbf{ 6.3} 			&&19.4  &\textbf{ 8.5} & 8.6  			&&20.0  &\textbf{ 5.9} & 6.5  	\\ 
    Gender &  20 		&&16.2  &10.0  &\textbf{ 7.8} 			&&13.2  & 7.1  &\textbf{ 5.1} 			&&14.1  & 8.4  &\textbf{ 7.8} 			&&15.4  & 5.9  &\textbf{ 4.9} 	\\ 
           &  40 		&&11.6  & 9.2  &\textbf{ 6.6} 			&&11.4  & 8.4  &\textbf{ 4.5} 			&&11.1  & 7.7  &\textbf{ 5.9} 			&&11.1  & 6.1  &\textbf{ 3.8} 	\\ 
           & 100 		&& 7.1  & 6.5  &\textbf{ 5.4} 			&& 6.9  & 6.6  &\textbf{ 3.7} 			&& 7.0  & 6.1  &\textbf{ 4.8} 			&& 5.9  & 6.4  &\textbf{ 2.8} 	\\ 
           & 200 		&& 3.2  & 3.3  &\textbf{ 3.0} 			&& 4.0  & 4.0  &\textbf{ 2.9} 			&& 3.6  & 3.4  &\textbf{ 2.9} 			&& 4.0  & 4.0  &\textbf{ 2.2} 	\\ 
\midrule 
  Compas-R &  10 		&&21.1  &\textbf{ 2.9} & 4.2  			&&20.7  &\textbf{ 4.0} & 4.8  			&&20.3  &\textbf{ 1.4} & 2.4  			&&23.1  &\textbf{ 6.6} & 8.4  	\\ 
      Race &  20 		&&14.8  &\textbf{ 2.8} & 3.3  			&&15.2  & 3.9  &\textbf{ 3.8} 			&&15.8  &\textbf{ 2.0} & 2.5  			&&16.6  &\textbf{ 7.8} & 8.0  	\\ 
           &  40 		&&11.7  &\textbf{ 3.0} & 3.0  			&&12.1  & 3.9  &\textbf{ 3.6} 			&&11.6  &\textbf{ 2.0} & 2.0  			&&10.9  & 9.9  &\textbf{ 8.1} 	\\ 
           & 100 		&& 6.8  & 2.9  &\textbf{ 2.8} 			&& 7.4  & 3.7  &\textbf{ 3.4} 			&& 8.5  & 2.1  &\textbf{ 1.8} 			&& 7.9  & 7.7  &\textbf{ 6.0} 	\\ 
           &1000 		&& 2.0  &\textbf{ 1.5} & 1.6  			&& 1.9  &\textbf{ 1.6} & 1.7  			&& 1.9  & 1.3  &\textbf{ 1.2} 			&& 1.9  & 1.9  &\textbf{ 1.8} 	\\ 
\midrule 
  Compas-R &  10 		&&21.3  &\textbf{ 3.8} & 5.0  			&&22.0  &\textbf{ 3.4} & 3.8  			&&23.4  &\textbf{ 3.5} & 4.4  			&&25.4  &19.1  &\textbf{13.7} 	\\ 
    Gender &  20 		&&18.5  &\textbf{ 3.8} & 5.1  			&&18.4  &\textbf{ 3.3} & 4.0  			&&17.4  &\textbf{ 3.3} & 4.6  			&&21.4  &23.8  &\textbf{12.3} 	\\ 
           &  40 		&&12.2  &\textbf{ 3.4} & 4.0  			&&13.0  &\textbf{ 3.0} & 3.3  			&&13.7  &\textbf{ 2.8} & 3.6  			&&15.0  &23.8  &\textbf{ 9.5} 	\\ 
           & 100 		&& 8.8  &\textbf{ 3.2} & 3.3  			&& 9.1  & 2.7  &\textbf{ 2.6} 			&& 8.5  &\textbf{ 2.1} & 2.7  			&& 9.8  &15.5  &\textbf{ 8.0} 	\\ 
           &1000 		&& 2.0  & 1.7  &\textbf{ 1.4} 			&& 2.2  & 1.4  &\textbf{ 1.3} 			&& 2.4  & 1.6  &\textbf{ 1.4} 			&& 1.9  & 1.9  &\textbf{ 1.8} 	\\ 
\midrule 
 Compas-VR &  10 		&&17.4  & 4.0  &\textbf{ 4.0} 			&&15.6  &\textbf{ 4.4} & 4.4  			&&15.7  & 2.6  &\textbf{ 2.4} 			&&19.7  &\textbf{ 6.1} & 6.5  	\\ 
      Race &  20 		&&13.5  & 4.7  &\textbf{ 4.3} 			&&13.7  & 5.0  &\textbf{ 4.8} 			&&13.6  & 3.3  &\textbf{ 2.9} 			&&15.9  &10.7  &\textbf{ 6.5} 	\\ 
           &  40 		&& 9.6  & 4.5  &\textbf{ 3.8} 			&& 9.6  & 4.5  &\textbf{ 3.9} 			&& 9.9  & 3.1  &\textbf{ 2.4} 			&&11.1  & 8.8  &\textbf{ 5.5} 	\\ 
           & 100 		&& 5.6  & 3.6  &\textbf{ 3.1} 			&& 5.2  & 3.8  &\textbf{ 3.4} 			&& 6.2  & 2.6  &\textbf{ 2.0} 			&& 6.6  & 6.8  &\textbf{ 3.7} 	\\ 
           &1000 		&& 0.9  & 0.8  &\textbf{ 0.8} 			&& 0.9  &\textbf{ 0.8} & 0.8  			&& 0.9  & 0.8  &\textbf{ 0.8} 			&& 1.1  & 1.2  &\textbf{ 0.9} 	\\ 
\midrule 
 Compas-VR &  10 		&&17.2  & 5.6  &\textbf{ 5.4} 			&&16.8  & 5.7  &\textbf{ 5.3} 			&&19.0  &\textbf{ 5.8} & 6.3  			&&21.3  &18.9  &\textbf{ 9.8} 	\\ 
    Gender &  20 		&&13.3  & 5.4  &\textbf{ 5.1} 			&&14.1  & 5.4  &\textbf{ 4.9} 			&&14.0  &\textbf{ 5.7} & 6.2  			&&16.0  &28.2  &\textbf{ 8.7} 	\\ 
           &  40 		&& 9.3  & 5.1  &\textbf{ 4.7} 			&& 9.7  & 4.9  &\textbf{ 4.5} 			&&10.5  &\textbf{ 5.3} & 5.7  			&&12.4  &30.9  &\textbf{ 6.9} 	\\ 
           & 100 		&& 6.4  & 3.7  &\textbf{ 3.4} 			&& 5.9  & 3.5  &\textbf{ 3.1} 			&& 6.3  &\textbf{ 4.2} & 4.4  			&& 7.1  &18.5  &\textbf{ 4.5} 	\\ 
           &1000 		&& 1.0  &\textbf{ 0.8} & 0.9  			&& 1.0  &\textbf{ 0.9} & 0.9  			&& 0.9  &\textbf{ 0.9} & 1.0  			&& 1.4  & 0.9  &\textbf{ 0.9} 	\\ 
\midrule 
     Ricci &  10 		&&17.7  &16.1  &\textbf{14.6} 			&&14.4  &\textbf{ 7.5} & 7.9  			&&12.2  &\textbf{ 1.9} & 2.1  			&&13.1  & 1.7  &\textbf{ 1.6} 	\\ 
      Race &  20 		&&11.2  &11.8  &\textbf{ 9.8} 			&& 9.3  & 7.2  &\textbf{ 7.1} 			&& 8.5  &\textbf{ 1.5} & 1.5  			&& 9.5  &\textbf{ 2.0} & 2.1  	\\ 
           &  30 		&& 7.4  & 7.7  &\textbf{ 6.5} 			&& 5.8  & 5.1  &\textbf{ 4.6} 			&& 6.0  & 1.1  &\textbf{ 1.1} 			&& 6.4  &\textbf{ 1.9} & 2.0  	\\ 
\bottomrule
\end{tabular}
}
\label{tab:ablation}
\end{table}

\clearpage
\newpage
\addcontentsline{toc}{section}{Sensitivity Analysis}
\section*{Appendix: Sensitivity Analysis for Calibration Priors}
As described in Section 2.4, we use the beta calibration model to recalibrate a model score $s$ for the $g$-th group according to 
\begin{align*}
f( s; a_g , b_g, c_g ) = \frac{ 1 }{ 1 + e^{ - c_g - a_g \log s + b_g \log ( 1- s ) }} 
\end{align*}
where $a_g$, $b_g$, and $c_g$ are calibration parameters with $a_g,b_g \geq 0$. With $a_g=1, b_g=1, c_g=0$, $f(\cdot; 1, 1, 0)$ is an identity function.
We assume that the parameters from each individual group are sampled from a shared distribution:
\begin{align*}
\log a_g \sim \mathrm{N}( \mu_a , \sigma_a ),
\log b_g \sim \mathrm{N}( \mu_b , \sigma_b ), 
c_g \sim \mathrm{N}( \mu_c , \sigma_c )
\end{align*}
where $\pi = \{\mu_a, \sigma_a, \mu_b, \sigma_b, \mu_c, \sigma_c \}$ is the set of hyperparameters of the shared distributions. 
As discussed in the main paper, in our  experiments  we set the hyperparameters as 
\begin{align*} 
\mu_a \sim \mathrm{N}(0,.4),
\mu_b \sim \mathrm{N}(0,.4),
\mu_c \sim \mathrm{N}(0,2), 
\sigma_a \sim \mathrm{TN}(0,.15), 
\sigma_b \sim \mathrm{TN}(0,.15),
\sigma_c \sim \mathrm{TN}(0,.75)
\label{equation:prior}
\end{align*}
These prior distributions encode  a weak prior belief that the model scores are calibrated by placing the mode of $a_g, b_g$ and $c_g$ at 1, 1, and 0 respectively. 
We used exactly these prior settings in all our experiments across all datasets, all  groups, and all labeled and unlabeled dataset sizes, which already demonstrates to a certain extent the robustness of these settings.

In this Appendix we describe the results of a sensitivity analysis with respect to the variances in the prior above. We evaluate our proposed methodology over a range of settings for the variances,   multiplying the default values  with different values of $\alpha$, i.e.
\begin{align*} 
\mu_a \sim \mathrm{N}(0,.4\alpha), \sigma_a \sim \mathrm{TN}(0,.15\alpha)\\
\mu_b \sim \mathrm{N}(0,.4\alpha), \sigma_b \sim \mathrm{TN}(0,.15\alpha)\\
\mu_c \sim \mathrm{N}(0,2\alpha), \sigma_c \sim \mathrm{TN}(0,.75\alpha)
\end{align*}
with $\alpha$ ranging from 0.1 to 10. We reran our analysis, using the different variance settings, for the specific case of estimating  the change $\Delta$ in accuracy estimates for the Adult dataset grouped by the attribute ``race," for each of the four classification models in our study and with different amounts of labeled data.

Table~\ref{tab:hyperparams} shows the resulting MAE values as $\alpha$ is varied.  The results show that the Bayesian calibration (BC) model is robust to the settings of prior variances. Specifically, as $\alpha$ varies from 0.1 to 10  the MAE values with BC are almost always smaller than the ones obtained with BB, and there is a broad range of values $\alpha$ where the MAE values are close to their minimum The results also show that the BC method has less sensitivity to   $\alpha$ when the number of labeled examples $n_L$ is large, e.g. $n_L=1000$.

\begin{table}[!h]
\caption{{\bf MAE for $\Delta$  Accuracy Estimates} of the adult data grouped by attribute ``race," with different values of $n_L$. Shown are mean absolute error (MAE) values between estimates and true $\Delta$ across 100 runs of labeled samples of different sizes $n_L$ for different trained models (groups of columns). Estimation methods are BB (beta-binomial) and BC (Bayesian-calibration) with different values of $\alpha$ (rows). BB uses only labeled samples, and BC use both labeled samples and unlabeled data. 
}
\centering
\resizebox{\textwidth}{!}{%
\begin{tabular}{@{}ccccccccccccccccc@{}}
\toprule 
& \phantom{a} &  \multicolumn{3}{c}{Multi-layer Perceptron}  
& \phantom{a}&  \multicolumn{3}{c}{Logistic Regression}
& \phantom{a} & \multicolumn{3}{c}{Random Forest} 
& \phantom{a} & \multicolumn{3}{c}{Gaussian Naive Bayes}\\ 
\cmidrule{3-5} \cmidrule{7-9} \cmidrule{11-13} \cmidrule{15-17}
Method && 10&100&1000 && 10&100&1000 && 10&100&1000 && 10&100&1000 \\ \midrule
BB		&&18.52 &8.48 &2.46		&&18.74 &9.14 &2.30		&&18.24 &9.00 &2.12		&&18.88 &9.54 &2.32\\ 
BC, $\alpha$=0.1		&&2.63 &2.60 &2.27		&&2.46 &2.49 &2.13		&&2.87 &2.84 &2.43		&&4.67 &4.51 &0.78\\ 
BC, $\alpha$=0.2		&&2.63 &2.56 &2.08		&&2.46 &2.51 &2.06		&&2.85 &2.83 &2.09		&&4.63 &3.95 &0.82\\ 
BC, $\alpha$=0.3		&&2.60 &2.52 &1.88		&&2.42 &2.51 &1.95		&&2.85 &2.79 &1.86		&&4.44 &3.36 &0.97\\ 
BC, $\alpha$=0.4		&&2.49 &2.46 &1.74		&&2.41 &2.57 &1.90		&&2.74 &2.82 &1.70		&&4.25 &3.06 &1.11\\ 
BC, $\alpha$=0.5		&&2.49 &2.38 &1.71		&&2.44 &2.60 &1.82		&&2.82 &2.77 &1.65		&&4.01 &2.86 &1.43\\ 
BC, $\alpha$=0.6		&&2.47 &2.37 &1.62		&&2.55 &2.62 &1.75		&&2.82 &2.88 &1.60		&&3.81 &2.79 &1.46\\ 
BC, $\alpha$=0.7		&&2.61 &2.48 &1.51		&&2.36 &2.63 &1.70		&&2.90 &2.86 &1.54		&&3.54 &2.80 &1.50\\ 
BC, $\alpha$=0.8		&&2.86 &2.30 &1.47		&&2.52 &2.73 &1.63		&&2.87 &2.86 &1.46		&&3.51 &2.77 &1.60\\ 
BC, $\alpha$=0.9		&&2.93 &2.27 &1.43		&&2.44 &2.82 &1.64		&&2.87 &2.90 &1.46		&&3.14 &2.91 &1.58\\ 
BC, $\alpha$=1.0		&&3.05 &2.31 &1.50		&&2.71 &2.74 &1.57		&&2.99 &2.96 &1.42		&&3.31 &2.85 &1.68\\ 
BC, $\alpha$=1.1		&&3.14 &2.37 &1.45		&&2.65 &2.86 &1.55		&&2.90 &3.10 &1.40		&&3.25 &3.03 &1.65\\ 
BC, $\alpha$=1.2		&&3.11 &2.19 &1.49		&&2.73 &2.80 &1.52		&&3.27 &3.01 &1.39		&&3.20 &3.03 &1.68\\ 
BC, $\alpha$=1.3		&&3.48 &2.30 &1.51		&&2.91 &2.94 &1.54		&&3.11 &3.21 &1.39		&&3.15 &2.96 &1.71\\ 
BC, $\alpha$=1.4		&&3.76 &2.28 &1.47		&&3.17 &3.01 &1.51		&&3.26 &3.21 &1.30		&&3.48 &3.21 &1.75\\ 
BC, $\alpha$=1.5		&&3.67 &2.20 &1.49		&&3.12 &2.94 &1.51		&&3.46 &3.05 &1.34		&&3.23 &3.19 &1.66\\ 
BC, $\alpha$=1.6		&&4.06 &2.24 &1.45		&&3.26 &2.93 &1.47		&&3.56 &3.13 &1.33		&&3.48 &3.17 &1.69\\ 
BC, $\alpha$=1.7		&&4.02 &2.27 &1.46		&&3.46 &3.15 &1.46		&&3.75 &3.10 &1.27		&&3.43 &3.19 &1.74\\ 
BC, $\alpha$=1.8		&&4.35 &2.14 &1.42		&&3.36 &3.09 &1.50		&&3.76 &3.26 &1.29		&&3.67 &3.22 &1.81\\ 
BC, $\alpha$=1.9		&&4.35 &2.30 &1.48		&&3.48 &2.94 &1.42		&&3.54 &3.30 &1.28		&&3.82 &3.35 &1.84\\ 
BC, $\alpha$=2.0		&&4.69 &2.16 &1.44		&&3.87 &2.99 &1.54		&&3.91 &3.46 &1.21		&&3.83 &3.18 &1.81\\ 
BC, $\alpha$=5.0		&&8.11 &2.54 &1.63		&&6.31 &3.32 &1.53		&&5.32 &4.13 &1.31		&&5.25 &3.82 &2.13\\ 
BC, $\alpha$=10.0		&&10.39 &2.63 &1.63		&&7.18 &3.83 &1.70		&&7.19 &4.41 &1.42		&&6.32 &4.08 &2.33\\ 
\bottomrule
\end{tabular}
}
\label{tab:hyperparams}
\end{table}

\clearpage
\newpage
\addcontentsline{toc}{section}{Error Results with LLO Calibration}
\section*{Appendix: Error Results with LLO Calibration}
Our hierarchical Bayesian calibration approach can be adapted to use other parametric calibration methods. In addition to the beta calibration method described in the main paper, we also experimented with LLO (linear in log odds) calibration. 

Table~\ref{tab:llo_accuracy} below shows a direct comparison of the mean absolute error (MAE) rate for estimation of differences in accuracy between groups (same setup as Tables 2 and 3 in the main paper in terms of how MAE is computed). The results show that in general the MAE of the two calibration methods tends to be very similar (relative to the size of the BB and frequentist MAEs)  across different dataset-attribute combinations. different prediction models, and different $n_L$ values. 

\begin{table}[!h]
\caption{{\bf MAE for $\Delta$  Accuracy Estimates of LLO and BC}, with different $n_L$. Mean absolute error between estimates and true $\Delta$ across 100 runs of labeled samples of different sizes $n_L$ for different trained models (groups of columns) and 10 different dataset-group combinations (groups of rows). Estimation methods are BC (Bayesian-Calibration) and LLO (Linear in Log Odds Calibration). Both methods use both labeled samples and unlabeled data. Trained models are Multilayer Perceptron, Logistic Regression, Random Forests, and Gaussian NaiveBayes.
}
\centering
\resizebox{\textwidth}{!}{%
\begin{tabular}{@{}rrrccccccccccc@{}}
\toprule 
& 
& \phantom{a} &  \multicolumn{2}{c}{Multi-layer Perceptron}  
& \phantom{a}&  \multicolumn{2}{c}{Logistic Regression}
& \phantom{a} & \multicolumn{2}{c}{Random Forest} 
& \phantom{a} & \multicolumn{2}{c}{Gaussian Naive Bayes}\\ 
\cmidrule{4-5} \cmidrule{7-8} \cmidrule{10-11} \cmidrule{13-14}
Group & $n$ && BC  &LLO && BC  &LLO && BC  &LLO && BC  &LLO \\ \midrule
     Adult &  10 		&& 3.9 & 3.8 		&& 2.9 & 2.8 		&& 3.2 & 3.2 		&& 3.6 & 3.5 \\ 
      Race & 100 		&& 3.5 & 3.4 		&& 3.2 & 3.1 		&& 3.1 & 2.9 		&& 2.8 & 2.4 \\ 
           &1000 		&& 1.6 & 2.3 		&& 1.7 & 2.0 		&& 1.4 & 1.5 		&& 1.4 & 1.6 \\ 
\midrule 
     Adult &  10 		&& 5.1 & 5.1 		&& 2.2 & 2.3 		&& 4.8 & 4.7 		&& 5.4 & 5.0 \\ 
    Gender & 100 		&& 4.4 & 4.3 		&& 1.9 & 2.0 		&& 4.1 & 3.7 		&& 2.7 & 2.7 \\ 
           &1000 		&& 1.6 & 2.2 		&& 1.1 & 1.0 		&& 2.0 & 1.5 		&& 1.1 & 1.1 \\ 
\midrule 
      Bank &  10 		&& 2.5 & 2.3 		&& 1.4 & 1.2 		&& 1.0 & 0.9 		&& 1.7 & 1.7 \\ 
       Age & 100 		&& 2.0 & 2.0 		&& 1.2 & 1.2 		&& 0.9 & 0.9 		&& 1.1 & 1.2 \\ 
           &1000 		&& 1.1 & 1.2 		&& 0.7 & 0.7 		&& 0.5 & 0.5 		&& 0.8 & 0.9 \\ 
\midrule 
    German &  10 		&& 5.0 & 4.6 		&& 8.7 & 8.0 		&& 8.2 & 7.5 		&&11.5 &10.7 \\ 
       age & 100 		&& 3.9 & 4.1 		&& 3.8 & 4.7 		&& 4.3 & 4.0 		&& 4.2 & 6.0 \\ 
           & 200 		&& 3.1 & 3.9 		&& 3.3 & 4.2 		&& 3.3 & 3.1 		&& 3.5 & 6.0 \\ 
\midrule 
    German &  10 		&& 8.2 & 6.4 		&& 6.3 & 5.0 		&& 8.6 & 6.9 		&& 6.5 & 5.3 \\ 
    Gender & 100 		&& 5.4 & 5.1 		&& 3.7 & 3.6 		&& 4.8 & 4.5 		&& 2.8 & 3.1 \\ 
           & 200 		&& 3.0 & 3.4 		&& 2.9 & 2.8 		&& 2.9 & 3.1 		&& 2.2 & 2.9 \\ 
\midrule 
  Compas-R &  10 		&& 4.2 & 4.6 		&& 4.8 & 5.2 		&& 2.4 & 2.5 		&& 8.4 & 8.2 \\ 
      Race & 100 		&& 2.8 & 4.4 		&& 3.4 & 4.8 		&& 1.8 & 1.4 		&& 6.0 & 5.6 \\ 
           &1000 		&& 1.6 & 5.0 		&& 1.6 & 4.4 		&& 1.2 & 1.1 		&& 1.8 & 2.9 \\ 
\midrule 
  Compas-R &  10 		&& 5.0 & 4.3 		&& 3.8 & 3.9 		&& 4.4 & 4.1 		&&13.7 &13.0 \\ 
    Gender & 100 		&& 3.3 & 2.7 		&& 2.6 & 2.3 		&& 2.7 & 2.8 		&& 8.0 & 7.4 \\ 
           &1000 		&& 1.4 & 2.1 		&& 1.3 & 1.3 		&& 1.4 & 3.0 		&& 1.8 & 2.4 \\ 
\midrule 
 Compas-VR &  10 		&& 4.0 & 3.9 		&& 4.4 & 4.7 		&& 2.4 & 2.9 		&& 6.5 & 6.4 \\ 
      Race & 100 		&& 3.1 & 2.8 		&& 3.4 & 3.3 		&& 2.0 & 2.1 		&& 3.7 & 3.6 \\ 
           &1000 		&& 0.8 & 1.5 		&& 0.8 & 0.8 		&& 0.8 & 2.5 		&& 0.9 & 1.8 \\ 
\midrule 
 Compas-VR &  10 		&& 5.4 & 4.8 		&& 5.3 & 5.2 		&& 6.3 & 8.2 		&& 9.8 & 9.0 \\ 
    Gender & 100 		&& 3.4 & 3.0 		&& 3.1 & 3.3 		&& 4.4 & 5.4 		&& 4.5 & 4.2 \\ 
           &1000 		&& 0.9 & 1.2 		&& 0.9 & 1.5 		&& 1.0 & 1.7 		&& 0.9 & 0.9 \\ 
\midrule 
     Ricci &  10 		&&14.6 &14.2 		&& 7.9 & 8.1 		&& 2.1 & 2.0 		&& 1.6 & 2.1 \\ 
      Race &  20 		&& 9.8 &13.6 		&& 7.1 & 6.6 		&& 1.5 & 1.6 		&& 2.1 & 2.5 \\ 
           &  30 		&& 6.5 &12.1 		&& 4.6 & 4.2 		&& 1.1 & 1.4 		&& 2.0 & 2.3 \\ 
\bottomrule
\end{tabular}
\label{tab:llo_accuracy}
}
\end{table}

\clearpage
\newpage
\addcontentsline{toc}{section}{Proof of Lemma 2.1}
\section*{Appendix: Proof of Lemma 2.1}

\begin{customlemma}{2.1}
Given a prediction model $M$ and score distribution $P(s)$,
let $f_g(s;\phi_g): [0,1] \rightarrow [0,1]$ denote the calibration model for group $g$;
let $f^*_g(s):[0,1] \rightarrow [0,1]$  be the optimal calibration function which maps 
$s = P_{M}(\hat{y}=1|g)$ 
to $P(y=1|g)$;
and $\Delta^*$ is the true value of the metric.
Then the absolute value of  expected estimation error w.r.t. $\phi$ can be bounded as:
$|\mathbb{E}_{\phi}\Delta - \Delta^*| \le \|\bar{f_0}-f^*_0\|_1 + \|\bar{f_1}-f^*_1\|_1$, 
where $\bar{f_g}(s) = \mathbb{E}_{\phi_g} f_g(s;{\phi_g}), \forall s\in[0,1]$, 
and $\|\cdot\|_1$ is the expected $L^1$ distance  w.r.t. $P(s|g)$. 
\label{lemma:delta_bias}
\end{customlemma}
\begin{proof}
\begin{align*}
   |\mathbb{E}_{\phi}\Delta - \Delta^*| 
   & =  |(\mathbb{E}_{\phi_1}\theta_1 - \mathbb{E}_{\phi_0}\theta_0) - (\theta^*_1 - \theta^*_0)|\\
   & \le  |\mathbb{E}_{\phi_0}\theta_0 - \theta^*_0| + |\mathbb{E}_{\phi_1}\theta_ 1- \theta^*_1| 
   &&\text{(triangle inequality)}\\
   & = \|\bar{f_0}-f^*_0\|_1 + \|\bar{f_1}-f^*_1\|_1
   &&\text{(Lemma~\ref{lemma:theta_bias})}
\end{align*}
\end{proof}

 \begin{customlemma}{2.2}
Given a prediction model $M$ and score distribution $P(s)$,
let $f_g(s;\phi_g): [0,1] \rightarrow [0,1]$ denote the calibration model for group $g$;
let $f^*_g(s):[0,1] \rightarrow [0,1]$  be the optimal calibration function which maps 
$s = P_{M}(\hat{y}=1|g)$ 
to $P(y=1|g)$;
and $\theta^*$ is the true value of the accuracy.
Then the absolute value of  expected estimation error w.r.t. $\phi$ can be bounded as:
$|\mathbb{E}_{\phi}\theta_g - \theta_g^*| \le \|\bar{f_g}-f^*_g\|_1$, 
where $\bar{f_g}(s) = \mathbb{E}_{\phi_g} f_g (s;{\phi_g}), \forall s\in[0,1]$, 
and $\|\cdot\|_1$ is the expected $L^1$ distance  w.r.t. $P(s|g)$.
 \label{lemma:theta_bias}
\end{customlemma}
 \begin{proof} 
 \begin{align*}
 \theta^*_g &= P(y=0,\hat{y}=0|g) + P(y=1,\hat{y}=1|g)\\
&= \int_{s < 0.5}P(y=0|s)P(s|g) ds +  \int_{s >= 0.5}P(y=1|s)P(s|g) ds\\
&= \int_{s < 0.5}(1-f^*(s))P(s|g) ds +  \int_{s >= 0.5}f^*(s)P(s|g) ds
 \end{align*}
Similarly, our method makes prediction about groupwise accuracy with calibrated scores given P($\phi$):
 \begin{align*}
\mathbb{E}_{\phi_g}\theta_g
&= \mathbb{E}_{\phi_g} \int_{s < 0.5}(1-f_g(s;\phi))P(s|g) ds +  \int_{s \ge 0.5}f_g(s;\phi)P(s|g) ds\\
&= \int_{s < 0.5}(1-\mathbb{E}_{\phi}f_g(s;\phi))P(s|g) ds +  \int_{s >= 0.5}\mathbb{E}_{\phi}f_g(s;\phi)P(s|g) ds\\
&= \int_{s < 0.5}(1-\bar{f_g}(s))P(s|g) ds +  \int_{s >= 0.5}\bar{f_g}(s)P(s|g) ds
 \end{align*} 
Then the absolute estimation bias of estimator $\mathbb{E}_{\phi \in \Phi}\theta_\phi$ is:
 \begin{align*}
 |\mathbb{E}_{\phi}\theta_g - \theta_g^*|
 &= \abs{\int_{s < 0.5}(\bar{f}(s)-f^*(s))P(s|g) ds +  \int_{s >= 0.5}(f^*(s)-\bar{f}(s))P(s|g) ds}\\
 &\leq \int_{s < 0.5}\abs{\bar{f}(s)-f^*(s)}P(s|g) ds +  \int_{s >= 0.5}\abs{f^*(s)-\bar{f}(s)}P(s|g) ds\\
  &= \int_s \abs{\bar{f}(s)-f^*(s)}P(s|g) ds\\
  &=\|\bar{f}-f^*\|_1
 \end{align*}
 \end{proof}
 
\end{document}